\documentclass[11pt,twoside,a4paper,openright]{book}
\usepackage{graphicx}
\graphicspath{ {./images/} }
\usepackage{wrapfig}

\usepackage[utf8]{inputenc}
\usepackage[T1]{fontenc}
\usepackage[english]{babel}

\usepackage{graphicx}

\usepackage[margin=2.9cm,inner=3.4cm]{geometry} 
\usepackage[nottoc,numbib]{tocbibind}
\usepackage[toc,page]{appendix}
\usepackage{tocbibind}
\usepackage{enumitem}
\usepackage{url}
\usepackage{cite}
\usepackage {xcolor}
\usepackage{indentfirst}
\usepackage{amsmath}
\usepackage{mathtools}
\usepackage{lscape}
\usepackage{caption} 
\usepackage{listings}
\usepackage{color}
\usepackage{float}
\definecolor{dkgreen}{rgb}{0,0.6,0}
\definecolor{gray}{rgb}{0.5,0.5,0.5}
\definecolor{mauve}{rgb}{0.58,0,0.82}

\usepackage{breakurl}
\usepackage[breaklinks]{hyperref}
\begin{document}
%================================================================
%================================================================
\frontmatter
%================================================================
%================================================================

%---COVER--------------------------------------------------------
\newcommand{\HRule}{\rule{\linewidth}{0.5mm}}

\begin{titlepage}

\begin{center}

% Upper part of the page
\includegraphics{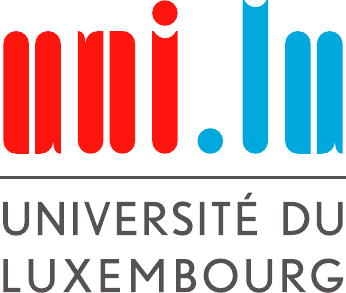}
%\unilulogo[1]
%\hfill
%\micslogo[1]

\vspace{1cm}

%\textsc{\LARGE University of Luxembourg}\\[1.0cm]
\textsc{\Large Faculty of Science, Technology and Communication}\\[1.0cm]

\vspace{1cm}

% Title
\HRule \\[0.4cm]
{\huge \bfseries Holographic Visualisation of Radiology Data and Automated Machine Learning-\\ based Medical Image Segmentation}\\
\HRule \\[1.5cm]

\begin{minipage}{0.8\textwidth}
\begin{center}
{\Large Thesis Submitted in Partial Fulfillment of the Requirements
for the Degree of Master in Information and Computer Sciences}
\end{center}
\end{minipage}

% use Prof. and Ass. Prof. as abbreviations for Professor and Associate Professor
% check csc.uni.lu/members for title
\vspace{4cm}
\begin{minipage}[t]{0.4\textwidth}
\begin{flushleft} \large
% the student who wrote the thesis
\emph{Author:}\\
Lucian \textsc{TRESTIOREANU}
\end{flushleft}
\end{minipage}
\begin{minipage}[t]{0.4\textwidth}
\begin{flushright} \large
% supervisor and reviewer are defined in the mics regulations (see moodle)
\emph{Supervisor:} \\
Dr.~Radu \textsc{STATE} \\
\vspace{.5em}
\emph{Reviewer:} \\
Prof. ~Dr. ~Christoph \textsc{SCHOMMER} \\
\vspace{.5em}
% advisor is a person that helps in the daily supervision
\emph{Advisor:} \\
Patrick \textsc{Glauner}
\end{flushright}
\end{minipage}

\vfill

% Bottom of the page
% month of submission
{\large August 2018}

\end{center}

\end{titlepage}

%---ABSTRACT-----------------------------------------------------
%\chapter*{Abstract}
\cleardoublepage %to fix page number in table of contents
\phantomsection  %to fix page number in table of contents
\addcontentsline{toc}{chapter}{Abstract}
\chapter*{\centering Abstract}

Within this thesis we propose a platform for combining Augmented Reality (AR) hardware with machine learning in a user-oriented pipeline, offering to the medical staff an intuitive 3D visualization of volumetric Computed Tomography (CT) and Magnetic Resonance Imaging (MRI) medical image segmentations inside the AR headset, that does not need human intervention for loading, processing and segmentation of medical images. The AR visualization, based on Microsoft HoloLens, employs a modular and thus scalable frontend-backend architecture for real-time visualizations on multiple AR headsets. \newline 
	\indent As Convolutional Neural Networks (CNNs) have lastly demonstrated superior performance for the machine learning task of image semantic segmentation, the pipeline also includes a fully automated CNN algorithm for the segmentation of the liver from CT scans. The model is based on the Deep Retinal Image Understanding (DRIU) model which is a Fully Convolutional Network with side outputs from feature maps with different resolution, extracted at different stages of the network. The algorithm is 2.5D which means that the input is a set of consecutive scan slices. The experiments have been performed on the Liver Tumor Segmentation Challenge (LiTS) dataset for liver segmentation and demonstrated good results and flexibility. \newline
    \indent While multiple approaches exist in the domain, only few of them have focused on overcoming the practical aspects which still largely hold this technology away from the operating rooms. In line with this, we also are next planning an evaluation from medical doctors and radiologists in a real-world environment.

%---ACKNOWLEDGEMENTS---------------------------------------------
\cleardoublepage %to fix page number in table of contents
\phantomsection  %to fix page number in table of contents
\addcontentsline{toc}{chapter}{Acknowledgements}
\chapter*{\centering Acknowledgements}

Firstly, I would like to thank Dr. Radu State for accepting me to join the SEDAN Lab for my master's thesis. I gained important knowledge and skills since being part of the project. I want to specially thank Mr. Patrick Glauner for his continuous support in all aspects during this whole time, and Dr. Jorge Meira for the advice during the project. I am happy I had the chance to work with you both and I learned a lot from the conversations we had along this semester. Thanks also to Mr. Max Gindt for the insights into practical medicine aspects and Mr. Martin Lehmann for his help with Unity. Finally, I want to thank my family, and especially my wife Lucia, who has always supported me and made this master possible. As well, I want to thank everyone I met during this master's program and have contributed to my development in computer science!

%--TABLE OF CONTENTS--------------------------------------------------
\cleardoublepage %to fix page number in table of contents
\phantomsection  %to fix page number in table of contents
\addcontentsline{toc}{chapter}{Contents} % add toc to toc
\renewcommand{\contentsname}{Table of Contents} % add toc to toc
\tableofcontents 
%---FIGURES TABLES LIST-----------------------------------------
\listoffigures
\listoftables
%=================================================================
%===============  MAIN CONTENT ==================================
%================================================================
%================================================================
\mainmatter
%%%%%%%%%%%%%%%%%%%%%%%%%%%%%%
% CHAPTER 1 - INTRODUCTION
%%%%%%%%%%%%%%%%%%%%%%%%%%%%%%%%%%
\chapter{Introduction}
Contemporary radiology is constantly evolving, giving us sharper and deeper medical imaging, from the initial capturing of Contrast Tomography (CT) images through to the handcrafted analysis. Organ and lesion segmentation helps doctors accurately diagnose disease, plan surgical interventions and the treatment of patients. Generally, the doctors are using CT and Magnetic Resonance Imaging (MRI) images segmented by manual or semi-automatic techniques.

	\section{Motivation and contributions}
 Despite some advancements in the domain, the current pipeline used by the practitioners is not automated and the quality of the final results, which are key to pre-operative decision making, depends on the resources available for segmentation and ad-hoc manipulation. In the recent years, automatic solutions which are less time-consuming and do not rely on the human factor have been investigated. Also, the volumetric image, three dimensional in nature, is viewed on two-dimensional displays. As such, a mass of information stays hidden from the viewer and the cognitive effort required to interpret the results is higher. \newline
 	\indent Today there are different approaches to present the 3D information provided by the different medical investigations in a 3D manner, but few of them address aspects of practicality and ergonomics such as the ones offered by AR and more specifically HoloLens: surroundings awareness and interaction using mechanisms unbound by traditional input hardware, which opens the possibility of using this technology in the operations theaters with strict hygiene requirements.
    Besides aspects like almost real-time access to segmented data, this work tries to leverage practical HoloLens advantages to create a valuable product for real-life and real-time medical usage. \newline \newline
    \indent More specifically, our project strives to address aspects like:
    \begin{itemize}[noitemsep]
            \item Being able to access the information (3D medical image and patient data) while during the situation (surgery, medical emergency),
            \item Information sharing (multiple users working in the same time on the same data, marking and pointing), 
            \item Strict hygiene rules in the operations rooms - the ability to control the device without having to touch any dedicated hardware,
            \item Surroundings awareness (be able to see the real-world environment in the same time with the medical data).
    \end{itemize}
     \indent \indent The main challenge in using HoloLens is the inherent limited rendering power associated to any mobile device, worsened by the high GPU demands required by the rendering of the 3D medical volumetric images (CTs and MRIs), which we address by using dedicated external hardware.
%=================================
		\subsection{Virtual and augmented reality head-mounted displays}
        \begin{wrapfigure}{r}{0.4\textwidth}
        \begin{minipage}{0.4\textwidth}
       		\includegraphics[width=1\textwidth]{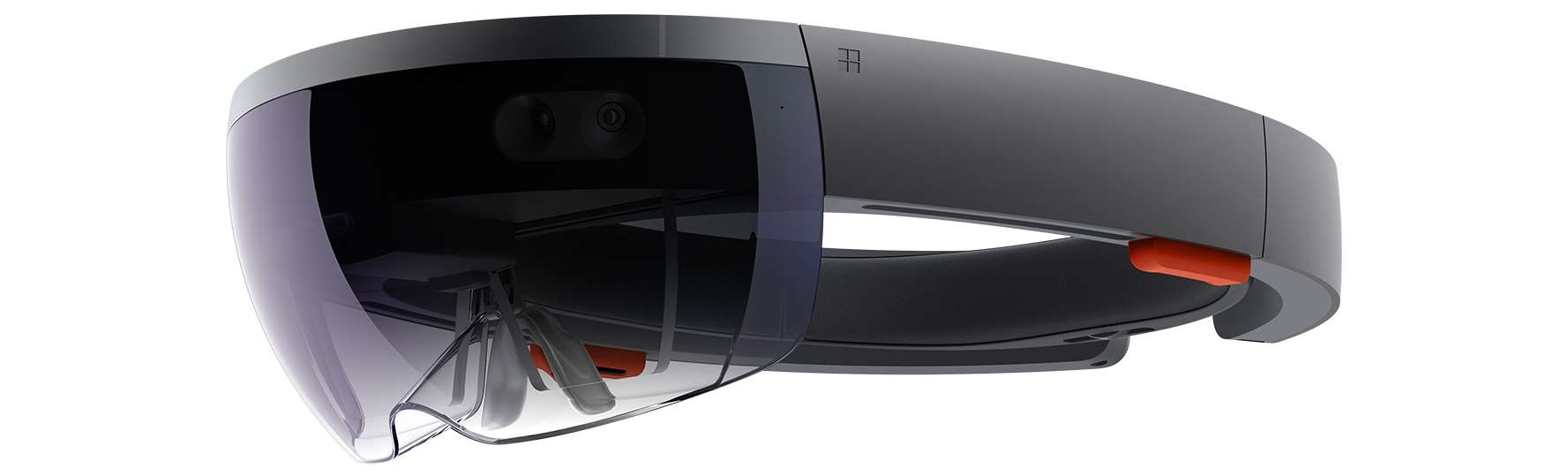}
			\caption[Microsoft HoloLens]{Microsoft HoloLens.\footnote{Source: \url{http://www.microsoft.com/en-us/hololens}}}
            \label{fig:holo}
        \end{minipage}
		\end{wrapfigure}
        As progress has been made on virtual reality (VR) and augmented reality (AR) devices, current headsets have demonstrated higher quality and lower prices. The most popular AR platform, Microsoft HoloLens \cite {max1}, presented in Figure \ref{fig:holo} , offers an advanced development and experimental platform.
        VR and AR devices offer 360$^{\circ}$ vision as well as haptics and interactions using mechanisms unbound by traditional machine-human interaction devices (like keyboards, pointer devices, controllers).
        \subsection{Use of VR and AR devices in the medical area}
        With the rapid progress of Graphics Processing Units (GPUs), the combined possibilities of AR or VR devices, and stronger GPUs, enable opportunities also in the medical area, where they can be used for the cognition of, and interaction with medical imaging not only during analysis but also during surgical interventions. Thanks to the GPUs massive computational power, heavy processing can be offloaded to a dedicated backend in order to be light on the low-resource headsets, e.g the HoloLens headset. \newline \indent
        Preoperative as well as intraoperative use of VR or AR models aiding decision-making has been proven to provide support to medical staff in diverse settings, inter alia in support of laparoscopic (or minimally invasive) surgery \cite{max2,max3} or by fusing endoscopic vision with MRI images \cite{max4}, which leads more and more research institutions into the area \cite {max5}. \newline \indent
       \subsection{Computer vision: the machine learning algorithms}
         \begin{wrapfigure}{r}{0.54\textwidth}
       		\includegraphics[width=0.54\textwidth]{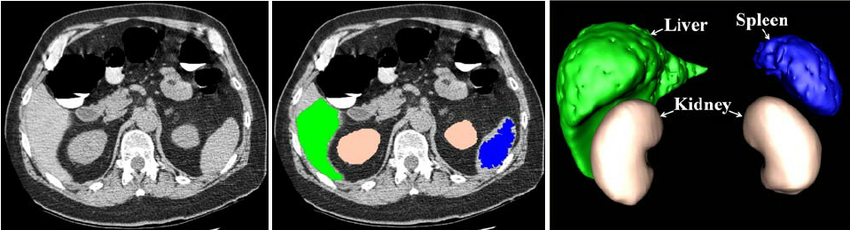}
			\caption[Medical image segmentation]{Medical image segmentation \cite{MIseg}.}
            \label{fig:segmentation}					
		\end{wrapfigure}
        Segmentation of images is the process of partitioning the image into semantically meaningful parts and classifying each part into pre-determined classes. An example is depicted in Figure~\ref{fig:segmentation}. As segmentation of medical images is a time consuming manual process, involving machine learning into this use-case has gained a high interest in the last years. \newline \indent 
        Generally, the automatic liver segmentation is challenging for reasons like noise in the CT scans, lesions variability, and low contrast of image elements such as liver, lesions and other organs. At present, the state of the art in solving these challenges has been achieved by Convolutional Neural Networks (CNNs) based approaches.
        \newline \indent 
        The algorithm we use for segmentation, adapts DRIU - Deep Retinal Image Understanding \cite{m24}, for the task of liver segmentation. This method has already had success with segmenting the eye's blood vessels and optical disk, as well as for One Shot Video Objects Segmentation (OSVOS) \cite{m2}. \newline 
        \indent Some important aspects of the algorithm we are using, are:
        \begin{itemize}[noitemsep]
            \item For solving the data imbalance, as explained in Section ~\ref{subsubsection:bce}, a weighting of the Binary Cross Entropy (BCE) loss used for training has been introduced.
            \item The employed framework is Tensorflow [45], starting from OSVOS' [3] open source code.
			\item The network is 2.5D, i.e. the input is a sequence of 3 consecutive CT scan slices, which improves the performance versus using a single slice.
			\item A 3D Conditional Random Field post-processing is also introduced in order to obtain spatial coherence for the predictions.
       \end{itemize}
    \indent  \indent This model has already demonstrated good performance during a liver segmentation dedicated competition, and also its generality is demonstrated on different anatomical structures from the Visceral dataset.
     %-----------------------
      \subsection {Contribution}
      %-----------------------
      Because of the high regulatory and scientific requirements of any medical device allowed to enter the operation rooms we decided to evaluate the best existing solutions and technologies that could be leveraged to achieve the stated goals. As a 3D medical image visualization solution we chose the HoloLens headset, and we chose to make use of the "3D Toolkit" project actively developed by the Microsoft Catalyst Partner Team \cite{catalyst} as the solution to power the external image rendering. \newline 
      \indent In regards to the liver segmentation requirement, the dataset we used comprises of 200 liver scans downloaded from the LiTS competition \cite{LITS}. The CNN liver segmentation model was made public at the same competition \cite{miriam}. \newline
        \begin{figure}[h!]
       		\includegraphics[width=1\textwidth]{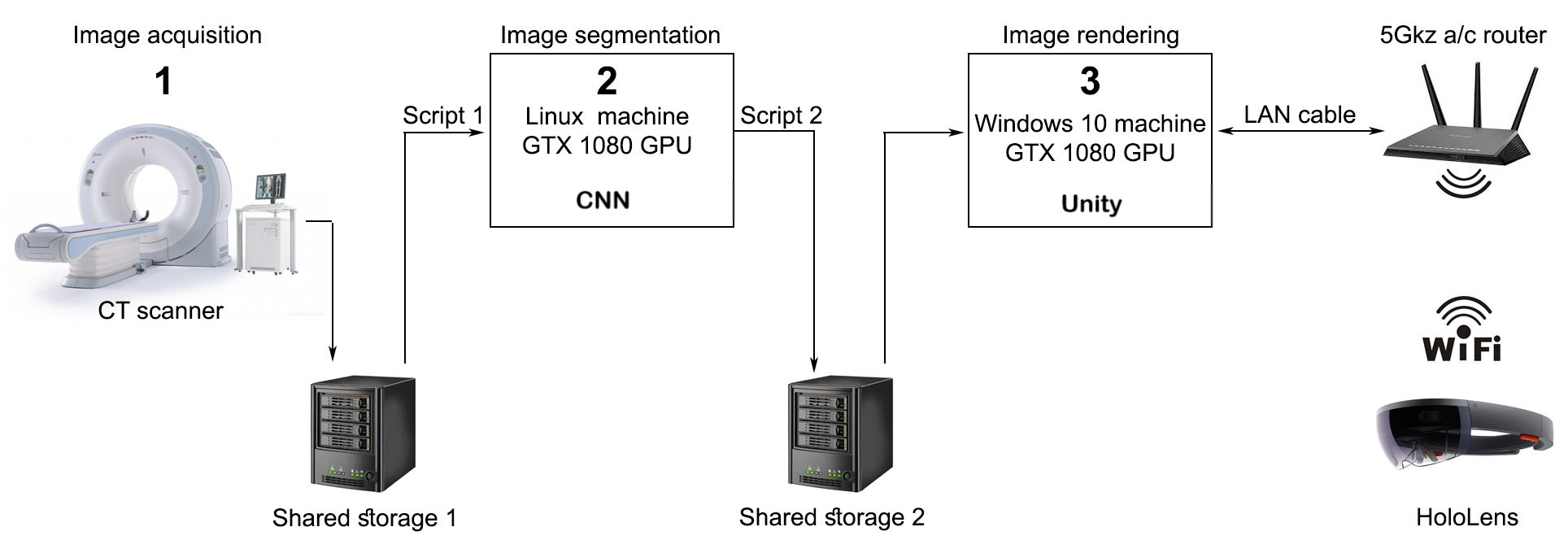}
			\caption[General system overview]{General system overview.}
            \label{fig: genoverview}					
		\end{figure}
      \indent In Figure \ref{fig: genoverview}, we present a general overview of the pipeline we propose. The CT scans acquired by the \textit{CT scanner} [Figure \ref{fig: genoverview}(1)] are stored in \textit{Shared storage 1}. \textit{Script 1}, running on the \textit{Linux machine} [Figure \ref{fig: genoverview} (2)], continuously monitors \textit{Shared storage 1}, and queues a new liver segmentation job as soon as it detects a new CT scan. \textit{Script 2}, running also on the \textit{Linux machine}, manages the new segmentation results: as soon as a new segmentation result is available it copies it in DICOM format on \textit{Shared storage 2} for classic visualization, starts the file conversion and copying procedure towards the Unity engine format, and optionally sends an update email to interested parties that a new segmentation result is available on \textit{Shared storage 2}. \textit{Shared storage 1} and \textit{Shared storage 2} can use the same physical hardware. Further, a 3D volumetric image rendering server based on the Unity game engine and running on the \textit{Windows 10 machine} [Figure \ref{fig: genoverview} (3)] powered by one or more high-end NVIDIA GPUs connected in SLI, renders the 3D volumetric image liver segmentation, encodes the video frames and sends them, through the \textit{5Ghz 802.11ac router}, towards the \textit{HoloLens Client} for visualization and manipulation as a mixed reality 3D volumetric image. Through the UI, the user can browse, load or unload for visualization the liver segmentations already prepared in Unity format on \textit{Shared storage 2} by the \textit{Script 2} running on the \textit{Linux machine}. \newline 
      \indent The advantage of having two dedicated machines is twofold: as both the 3D visualization and the medical volumetric image segmentation using a CNN are resource-intensive, the visualization performance will not be affected by a (possibly overlapping) running segmentation (ensure no hardware resource sharing), and also, the machine learning works at its best on Linux. Our experiments have also shown that for best latency, the router should be a model capable of running in the 802.11ac WiFi standard at 5Ghz, and channel bandwidth should be set as large as possible (we used 80Mhz). \newline 
      \indent In Chapter 2 we review the State of the Art in the field, to detail in Chapter \ref{arvrintro} why we chose the Microsoft HoloLens for the 3D  visualization, and in the following Chapters we explain how we used the solutions stated above to achieve the desired outcomes. 
    %-----------------------    
	\section{Work Packages} 
    %----------------------
As already discussed, the overall goal of the project is to present the end-user (medical staff) with an intuitive and high-value 3D visualization inside their headsets, that does not need human intervention for the segmentation. The Microsoft HoloLens headset has been selected because, especially in situations where several surgeons and medical staff are interacting in stressful situations with strict hygiene constraints, the use of this lightweight AR device needing no controllers could be beneficial, as it could provide new ways of interacting with the imagery and analysis during the situation. 
    	\subsection{WP1}
The aim is to build a fully automated pipeline, using a modular and thus scalable frontend-backend architecture, that should deliver real-time visualizations to multiple AR or VR headsets. A number of elements are needed in this context:

          \begin{itemize}[noitemsep]
            \item A modular backend architecture deployed in a (CUDA based) GPU architecture;
            \item Input file types must include DICOM as well as ultrasound industry-standard files;
            \item Output files should be optimized for visualization on the Unity game engine;
            \item UI and volumetric (or equivalent) rendering in a Unity-based executable;
            \item The executable needs to include basic volume information of the region of interest, needed for medical decision making;
            \item Multiple headsets should be able to view the same 3D objects at the same time, with an option for shared cooperative visualization (a few clients seeing the same thing at the same time) and marking/pointing;
            \item The architecture needs to be low-latency and should offload a maximum of processing to the backend in order to be light on the low-resource headsets, in the case of the HoloLens headset.
          \end{itemize}
      
		\subsection{WP2}
        After the acquisition of CT or ultrasound images and their deposit on the backend, they should be segmented without human intervention. The inference process and the subsequent transformation/conversion of the file should be near-real-time (under 60 seconds). Significant progress has been achieved in the domain, through the usage of state of the art machine learning algorithms. For this, WP2 needs to achieve these complex targets:
        	\begin{itemize}[noitemsep]
            	\item Human equivalent ML-based image segmentation of an organ in CT imagery;
                \item Calculate the volumetric data of the segmented organ;
                \item Anatomical feature segmentation inside the organ (i.e.arteries) or region of interest;
                \item Clean-up of intraoperative ultra-sound imagery, identification of features based on the region of interest identified in the CT image.
            \end{itemize}
            
         The last point, possibly the most complex to implement, could increase the situational awareness of medical staff in intra-operative situations by simplifying the analysis of ultrasound imagery.
%$$$$$$$$$$$$$$$$$$$$$$$$$$$$$$$$$$$$$$$$$$$$$ 
%$$$$$$$$$$$  CAP. 2 - RELATED WORK $$$$$$$$$$$$$$$$$$$
%$$$$$$$$$$$$$$$$$$$$$$$$$$$$$$$$$$$$$$$$$$$$$
\chapter{State of the Art}
	This chapter contains a presentation and review of the current state of the art in the fields of machine learning, augmented reality and virtual reality, which is then narrowed down more specifically to their current development and utilization in the field of medical imaging.
	\section{Machine learning}
\begin{figure}[h!]
   		 	\begin{minipage}{\textwidth}
       			\includegraphics[width=1\textwidth]{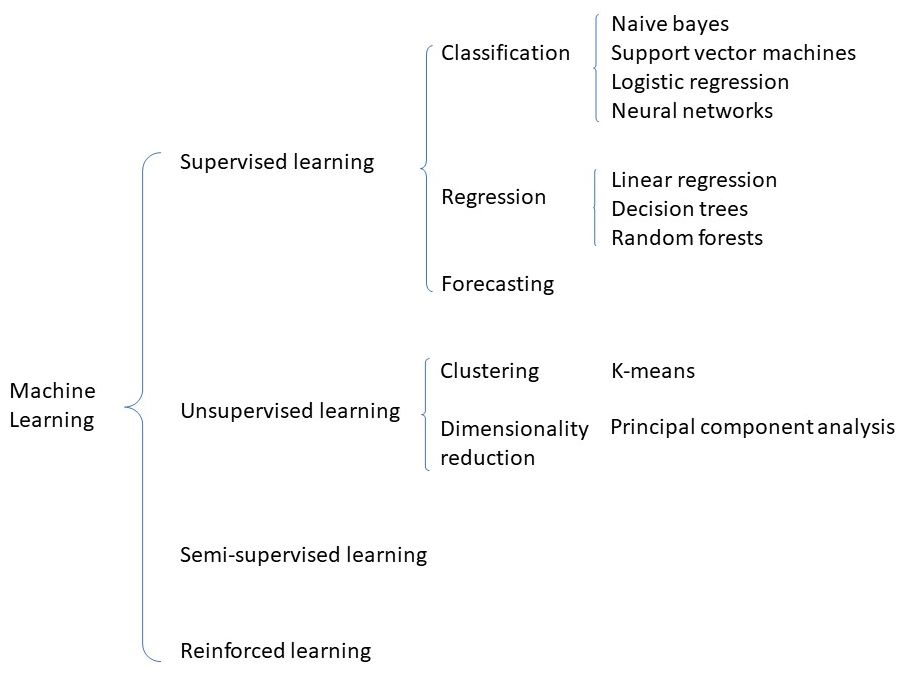}
				\caption[Machine learning algorithms]{Machine learning algorithms.}
           		\label{fig: ML}
			\end{minipage}
    	\end{figure}

     \indent The field of Machine Learning (ML) is part of the larger field of Artificial Intelligence (AI) \cite {glauner2017top}. The term "machine learning' was apparently first introduced by Mr. Arthur Samuels, an American scientist,  in 1959. He defined it as a “computer’s ability to learn without being explicitly programmed” \cite{samuel}. Machine learning receives data and processes it in order to make predictions. The machine learns as it interacts with new data, situations and as it checks its prediction against some given example outputs. More formally, ML algorithms learn a target function $f$ that best maps input variables $x$ to an output variable $y$: $y = f(x)$. \newline \indent
     There are four types of machine learning algorithms: supervised, unsupervised, semi-supervised, and reinforcement learning:
     \newline \indent In \textbf{supervised learning}, the machine is "taught by example". The algorithm is given a set of possible real-life inputs and outputs, and it should learn how to correctly answer a similar input, never seen before. While the human supervisor already knows the correct answers (has labeled the data set), from the examples it sees, the model has to make connections and patterns in order to be able to answer new queries. The model produces its own predictions and is corrected using the data labeling given by the human supervisor; the process continues until the model learns enough to reach the desired goals and performance. Ex.: The \textit{K-Nearest-Neighbor} algorithm estimates how likely a data point is to be a member of one group or another \cite{knn}. It checks the data points around a given point in order to determine to which group the given point is most likely to belong to. If the majority of the neighboring points are in the "circle" class, for example, it is likely that the point of interest is in the same class. In fact, "supervised learning" includes multiple categories:
     \begin{itemize}[noitemsep]
         \item Classification: In classification tasks, the algorithm should infer a conclusion from the examples given and further it should be able to determine to which categories the new inputs belong. For example, when deciding if a given transaction is legitimate, the algorithm should check the examples which it already has been provided with, and take a decision according to these. Ex.: \textit{Naive Bayes Classifier} \cite[p.~488]{AIMA}, based on Bayes’ theorem, classifies every value as independent of any other value. It allows predicting a class/category, based on a given set of features, using probability. \textit{Support Vector Machines} \cite{svm} analyze data used for classification and regression analysis. They filter data into categories, by providing a set of training examples, each marked as belonging to one or the other of the two categories. Afterwards, a model that assigns new values to one category or the other is built. \textit{Logistic regression} \cite{logreg} focuses on estimating the probability of an event to occur, based on the previous data provided. It is used to cover a binary dependent variable, where only two values, 0 or 1 are possible outcomes.
		\item Regression: In regression tasks, the algorithm should estimate the relationship between variables. Regression analysis studies the relation between one dependent variable and some set of changing variables, which makes it valuable for tasks such as prediction or forecasting. Ex.: \textit{Linear Regression} \cite{linregref}, which allows understanding the relationship between multiple continuous variables. \newline \indent
		Other examples of classification and regression are \textit{"Decision Trees"} \cite{dectreesref,dectreesref2} - which can be seen as a diagram with a tree structure that uses branching in order to represent all possible outcomes of a given problem, and \textit{"Random Forests"} \cite{randforref2, randforref3, randforref} which is an "ensemble learning" method, and combines multiple algorithms in order to generate better results for tasks such as classification or regression. Alone, each classifier is "weak", but put together, they are able to produce better results.
		\item Forecasting, which is the process of making predictions based on the past and present data. It is used often for trends analysis.
    \end {itemize}
    \indent \indent 
    In the case of \textbf{unsupervised learning}, the machine learning algorithm independently studies the data to identify patterns, without any answers or instructions provided. The algorithm identifies correlations and draws conclusions automatically, by analyzing the provided data. In the case of unsupervised learning, the computer algorithm interprets the data sets and processes the data as needed. The algorithm tries to organize the data in some way to describe its structure. This might involve data clustering or arranging it in a more organized way. As it continues to process more data, the algorithm improves its correct decision-making ability and becomes more refined. The following most common categories can be identified:
	\begin{itemize}[noitemsep]
         \item Clustering, which involves grouping sets of similar data (based on defined criteria). It’s useful for segmenting data into several groups and performing analysis on each data set to find patterns. Ex.: \textit{K-Means Clustering} \cite{kmeansref}, used to categorize unlabeled data, i.e. data without defined categories or groups. It works by finding groups within the data, with $K$ representing the number of groups. The algorithm performs successive iterations with the goal to assign each data point to one of the $K$ groups based on the features provided.
		\item Dimensionality reduction algorithms, such as the Principal Component Analysis (PCA) \cite{pca} reduce the number of variables being considered, in order to find some exact information required.
    \end{itemize}
    \indent \indent \textbf{Semi-supervised learning} is similar to supervised learning, with the difference that it makes use of both labeled and unlabeled data. Labeled data is information that has some meaningful tags attached, such that the algorithm can interpret it, while unlabeled data is missing such tags. This way, machine learning algorithms can learn to label data. \newline
    \indent \textbf{Reinforcement learning} (RL) draws inspiration from "behaviorist" psychology, and is concerned with how software should act in a given environment so as to maximize some reward. Using as input some parameters, output values and possible actions, the algorithm must try different options and possibilities and evaluate possible results, by means of defining some rules, in order to find the best solution. Reinforcement learning is based on the concept of learning by trial and error. It learns from experience and tries to adapt according to the given situation, in order to achieve the best possible result.\newline 
    \indent An overview of the classification of machine learning algorithms is illustrated in Figure \ref{fig: ML}. 
    \subsection{Artificial neural networks}
\begin{figure}[h!]
    		\centering
   		 	\begin{minipage}{0.8\textwidth}
       			\includegraphics[width=1\textwidth]{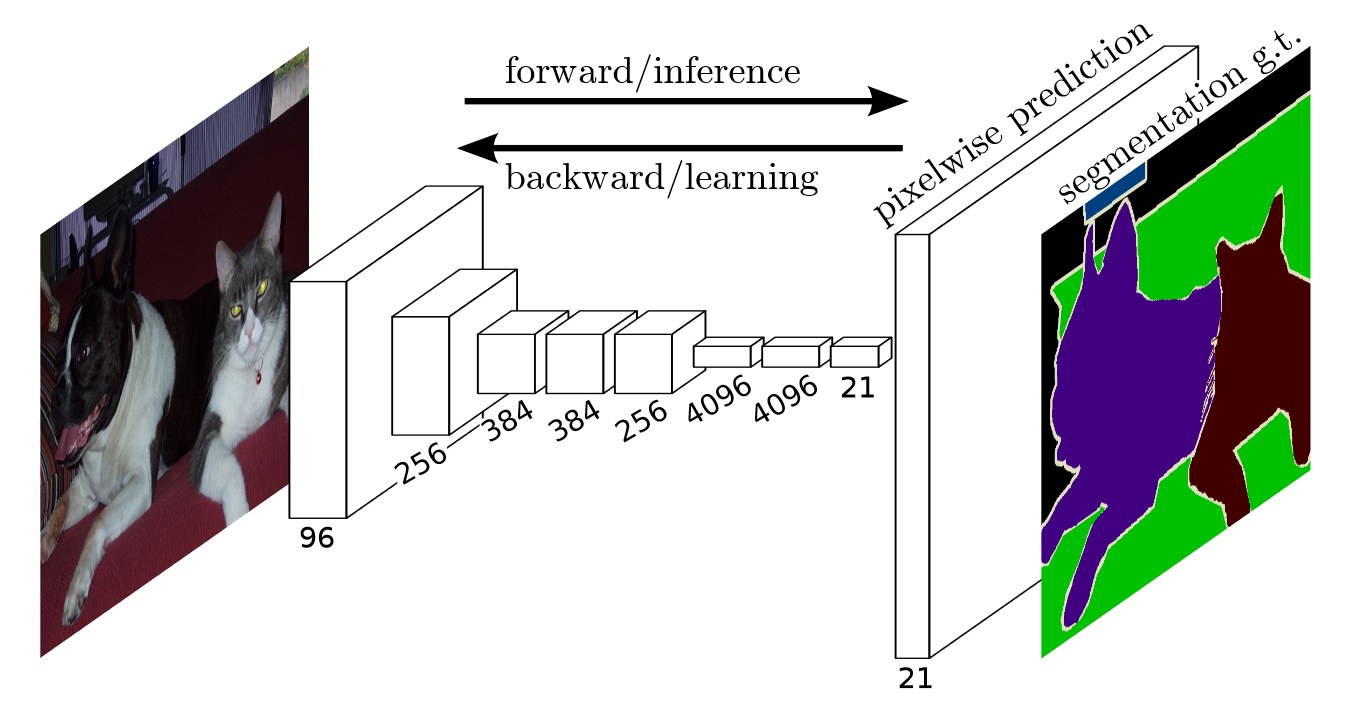}
				\caption[Fully connected networks]{Fully connected networks \cite{m21}.}
           		\label{fig: FCN}
			\end{minipage}
    	\end{figure}

    \textit{Artificial neural networks} (ANN) \cite{glauner2015comparison, glauner2015deep, glauner2016deep} are an important exponent of \textit{supervised learning}. They are made of "units" arranged in layers, each of which being connected to other layers on its sides. ANNs are inspired by biological systems, such as the brain, and are in fact a large number of interconnected processing elements, working together to solve some given problem. ANNs are learning by example or through experience, and are most useful for tasks such as modeling non-linear relationships in high-dimensional data or where the variables' relationship is very complex. \newline \indent There are different variants of ANNs, such as deep multilayer perceptron (MLP), convolutional neural network (CNN), recursive neural network (RNN), recurrent neural network (RNN), long short-term memory (LSTM), sequence-to-sequence model, and shallow neural networks. \newline
    \indent In the case of supervised learning, deep Convolutional Neural Networks (CNNs) have greatly increased in the recent years the accuracy of the computer vision algorithms for segmentation and detection. Description of features has been a long-standing issue in machine learning. It has become easier to train algorithms that can recognize optical features in highly diverse sets of images with acceptable error rates, often similar to human rates, or better. Starting with AlexNet \cite{m17}, the proposed solutions have improved. VGGNet \cite{m28} and ResNet \cite{m12} are between the best of today. VGGNet is built of convolutional stages with pooling layers which decrease the resolution of the feature maps and increase the neurons' receptive field. The deep network layers learn coarse features and the local features are learned by the beginning layers. In order to solve the gradient vanishing problem ResNet shortcuts each layer, creating thus a better gradient flow.
    For computer vision, excepting image classification, deep learning is also used for tasks like segmentation or boundary detection \cite{m23,m29}. Dense predictions solutions are using Fully Convolutional Networks \cite{m21}, as depicted in Figure \ref{fig: FCN}.
    
    \indent While CNNs used for Image Recognition include convolutional, pooling and fully connected layers, FCNs use just convolutional and pooling layers, and are thus able, with just one forward pass, to produce segmentation maps for all sizes of inputs.
    There are two main possibilities to output the initial resolution:
    \begin{itemize}[noitemsep]
         \item An expanding path consisting of deconvolutional or unpooling layers, which recovers spatial information by using features from different levels of the contracting path. As such, important low-level details can be preserved.
		\item Use of dilated convolutions \cite {m30} to increment the field of view, and thus preserving a higher features' resolution in the network.
    \end{itemize}
    
    %------------------
    \subsection{Machine learning for medical imaging}
     Under the machine learning umbrella there is a set of highly diverse algorithms. However, some techniques have proven to be more useful for medical imaging than others \cite{max6}. Some research cases have focused on techniques like \textit{clustering algorithms} (e.g. \textit{nearest neighbors}) or \textit{support vector machine} algorithms \cite {max7}, while others have investigated \textit{deep neural networks} \cite {max8, max9, max10, max11, max12, max14, max15, max16, max17, max18, max19, max20}. \newline
      \indent In regards to liver and lesion segmentation, lastly, most methods are based on \textit{statistical shape} and \textit{intensity distribution models} \cite {m14}. In \cite{liseg}, the authors argue that the statistical shape methods have the disadvantage that the resulting model may be too constraining if a patient’s liver shape is not adequately represented in the training data. This may result in limited application to pathological or post-operative livers. Other approaches rely on \textit{classifiers} and \textit{low-level segmentation} \cite {m15}, by using a first stage for organ detection and "active contours" in a second stage for segmenting the lesion. \textit{Support vector machines} (SVMs) and \textit{random forests} approaches, although having a higher discriminative power than the intensity based techniques, can lead to coarse segmentation and leakage \cite{liseg}. \newline \indent 
	Because of the flexibility and complexity of the learned features, neural networks, in their different flavors, empirically hold the most potential to infer high-level features in many computer vision tasks with acceptable error rates. Because they are robust to image variation, which allows building automatic segmentation, \textit{deep CNNs} have recently been proven successful for these tasks. This technique allows for the segmentation of heterogeneous livers obtained using different scanners and protocols in under 100 seconds \cite{liseg}. The initial segmentation of a region of interest (ROI), such as an organ or a structure like bone, in a medical image, is a use case investigated by a large number of papers but identifying pathologies or other features (i.e. the vascular system of an organ) in this segmented ROI is a second interesting use-case. \newline
    \indent DRIU \cite {m24}, the basis of the model presented here, was intended to segment the optical disc and blood vessels of the eye. The Fully Convolutional Network (FCN) used by DRIU has side outputs with supervision at different convolution stages\cite{miriam}. The output is obtained by combining the multi scale side outputs. \newline
    \indent In the medical domain, such algorithms have been used to detect anatomical features or lesions on every possible scale, from the microscopic level upwards. Some challenging characteristics for training CNNs like the imbalance of the labels for the data, require some modifications of the classic approaches though \cite{glauner2016large, glauner2017challenge}. \newline
    \indent The medical images, presented in the form of 3D volumes, can be processed using several methods: 
    \begin{itemize}[noitemsep]
        \item Using 2D FCNs: DRIU \cite{m24}, U-net \cite {m27} or DCAN \cite{m3}. They independently consider each slice of the volume, to provide in the end the segmentation of the 3D volume
by concatenating the 2D results. They have a disadvantage at spatial correlation on the $z$-axis.
		\item Implementing a 3D-convolutional network - to utilize the 3D characteristic of the data, or a hybrid one (2D and 3D convolutions) \cite {m19}. In \cite{cicek}, Çiçek et al. transform the 2D U-net architecture into a 3D-Unet which inputs 3D volumes and processes them with similar 3D operations like 3D convolutions, max-pooling, and up-convolutional layers. The 3D networks face the challenge of high computational costs.
        \item Combine several tri-planar schemes, as explained in \cite {m10}.  These apply three 2D convolutional networks based on orthogonal planes in order to classify the 3D pixel (voxel).
        \item Capture the volume information using Recurrent Neural Networks (RNNs) \cite {m4}. A 3D medical image is usually a sequence of 2D images, and RNNs are effective at processing sequential data.
    \end{itemize}
    \indent \indent Concerning liver segmentation, \cite {m8} builds 3D Fully Connected Networks and adds Conditional Random Fields as a post processing step, while \cite {m22} proposes 3D CNNs with a Graph Cut. Some approaches train two FCNs: the first processes the liver, while the second uses the liver mask for a better lesion segmentation \cite {m6}. Using the formulation by [16], they also add a 3D-CRF post-processing step in order to obtain spatial coherence in all the dimensions of the 3D input image. \newline \indent
        The best model from the 2017 LiTS challenge \cite {m11} trained from scratch a 2.5D DCNN having a set of adjacent channels as input, and using long and short range residual blocks connections. First, they train a network that outputs the approximate liver location and then they introduce 3D connected component labeling as a post-processing step.
        \newline \indent Deep learning (a 2D U-net architecture) together with a regular classifier (random forest) have also been used by \cite {m5}. They first perform the liver segmentation with an ensemble of CNNs. Second, the lesion segmentation is performed on a dedicated network followed by connected components. In the end a forest classifier filters the false positives. A residual network, which creates a cascaded architecture gradually improving on the segmentation from the previous step, has also been proposed by \cite {m1}. \newline
	\indent While the training period of deep neural networks for liver segmentation using a large set of images (e.g. for n=100, data size reaches 50Gb) can be time-intensive (depending on computational resources), the time necessary for the actual usage (inference) of the algorithm on an image that was not seen before by the network is nearly instant (typically under 60 seconds, depending on the model). The duration of the training period is a limiting factor for experimentation as it takes longer to try out different possibilities, but it can be radically shortened by using dedicated hardware (for instance clusters of GPUs).    
    \newline \newline
    \section{Medical imaging}
    In 2015, the United States alone spent 17\% of its gross domestic product (GDP) on healthcare, out of which  approximately 3\% of the GDP includes costs related to surgery \cite{m31}. As such, there is strong motivation to increase the efficiency in the operation rooms in order to both improve patient care and decrease cost. In this section we will review the state of the current equipment and visualization methods, to finally show that augmented reality (AR) and virtual reality (VR) can help increase efficiency and effectiveness, by improving pre-planning and intra-operative surgery [6]. \newline \indent
    
 	\subsection{Current medical imaging equipment}
\begin{figure}[h]
        	\centering
       		\includegraphics[width=0.6\textwidth]{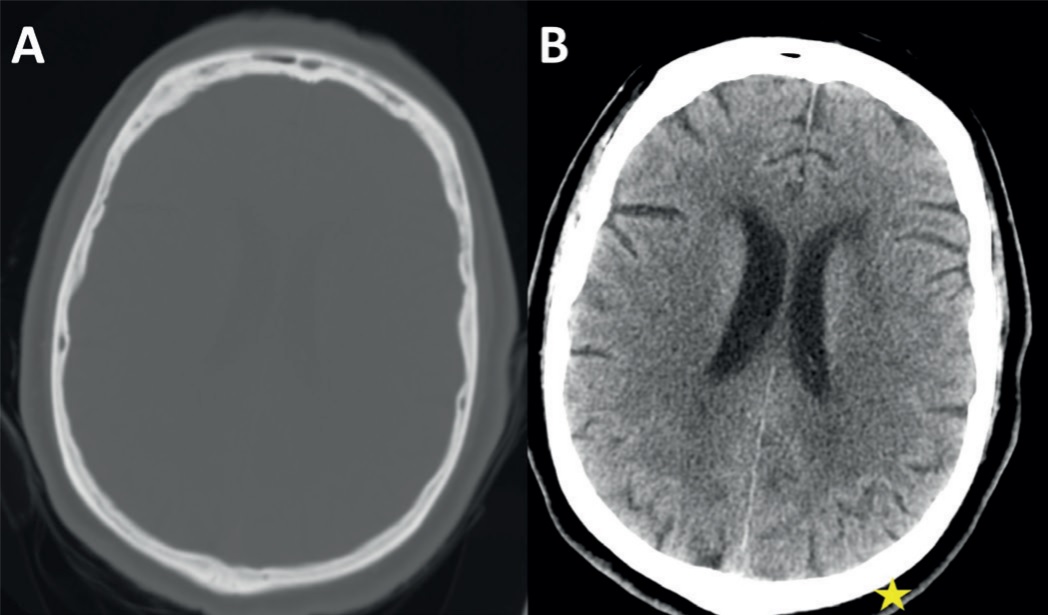}
			\caption[Axial head CT in “bone window” and “brain window”]{Axial head CT in “bone window” (A) and “brain window” (B) \cite{m31}.}
            \label{fig:CT scan}					
		\end{figure}

    The medical imaging equipment can generate 2D images (radiographs) as well as 3D volumetric images datasets. Examples of 3D imaging systems are computed tomography (CT), magnetic resonance imaging (MRI), and positron emission tomography (PET). \newline \indent
    In the case of the \textbf{CT scans}, the data is stored in "Digital Imaging and COmmunication in Medicine" (DICOM) file format where a regular matrix for a CT is 512 by 512 pixels. In a 2D plane, a pixel has one length in the $x$ direction and one length on the $y$ direction. By adding a third dimension to the pixel, a 3D volume object is created, which is called a voxel.
        Each pixel has a gray scale value called a Hounsfield Unit (HU), which in medical imaging is a function of tissue composition. For example the water's HU is zero, and soft tissue like the brain, kidney, muscle has an HU between 30 and 40. Bones can have an HU of 400 while air (less dense than water) has -1000. 
    \newline \indent The radiologists perform a task named “windowing and leveling” which means they set the window “level” and window “width”. An example is depicted in Figure~\ref{fig:CT scan}. The "window level” is the HU value for representing mid-grays. The “width” is the range of gray values where everything larger or smaller than the "width" is displayed in black or white.     	
     Also for the case of \textbf{magnetic resonance imaging (MRI)}, continuous planar images can be stacked and axial (head - toes), sagittal (left-right) and coronal (chest to back) reformats can be reconstructed. An MRI scan is similar to a CT scan in matrix size and pixel gray scale values. Unlike CT, MRI can produce very good contrast resolution between tissues of similar density and thus can diagnose some injuries which remain hidden on CT scans. MRI scans take longer than CTs to acquire, and sometimes the patient's breathing can make it difficult to acquire a good quality scan.
     \subsection{Current medical data visualization methods}
\begin{figure}[h]
       		\includegraphics[width=1\textwidth]{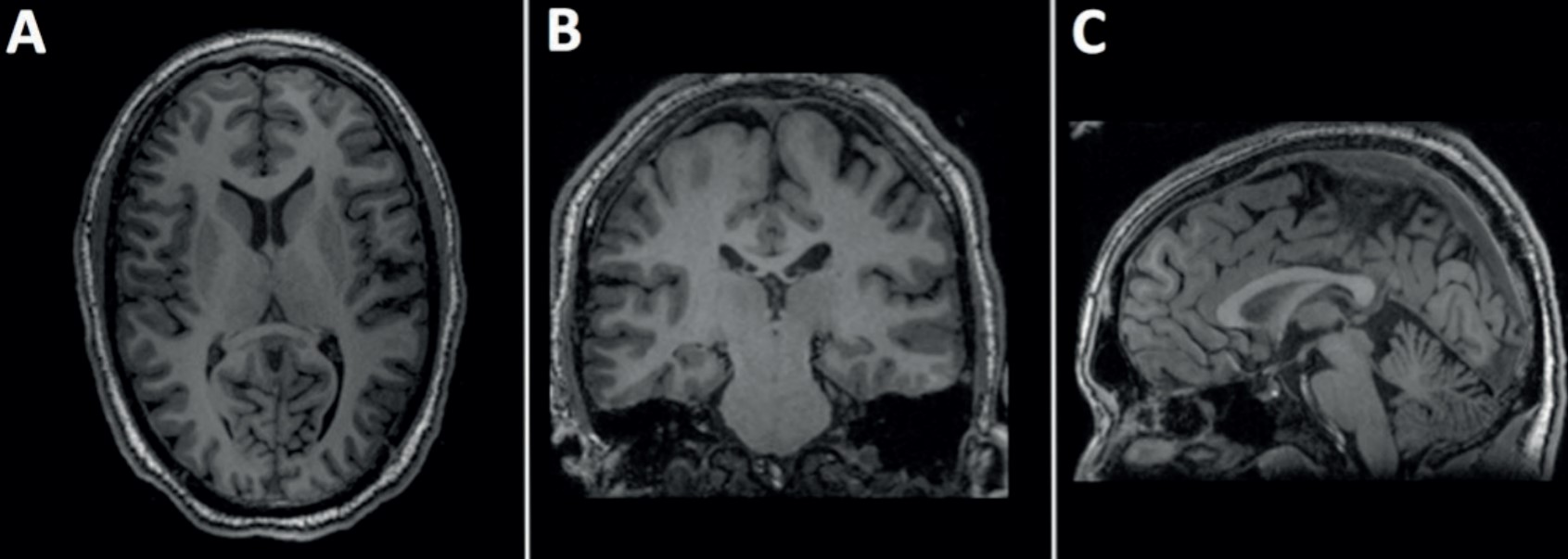}
			\caption[MRI brain scan: axial, coronal and sagittal images]{MRI brain scan: axial (A), coronal (B) and sagittal (C) images \cite{m31}.}
            \label{fig: Conventional viewing of the volumetric data}					
	\end{figure}

     Current methods for visualization of medical data include the \textbf{conventional viewing of the volumetric data} which is a slice-by-slice viewing method for axial, sagittal and coronal imaging planes or sometimes oblique reformats \cite{m31}, as shown in Figure \ref{fig: Conventional viewing of the volumetric data}. 
    \newline \indent Challenges involved by conventional viewing of the volumetric data include the challenge of information overload which comes from the volume of the generated datasets. A second challenge is detecting small lesions, where a very detailed slice by slice analysis will take considerable time. Also, there is the challenge of mentally building a 3D image by reviewing slices \cite{missedcancer1,missedcancer2,3dplan}. \newline \newline 
    \indent \textbf{Surface rendering} is the first 3D rendering method used for displaying 3D medical images \cite{m31,Douglas2}. The surfaces are displayed using segmentation techniques such as thresholding which helps to select only the designated pixels. A virtual light provides surface shading. This method allows presenting only a single surface, which results in the relative advantage of no overlapping tissues. This becomes a limitation, though, when trying to understand the relationships between multiple organ systems. Also, many organs have similar density with the surroundings and as such, they can be difficult to separate. Lastly, true depth perception can not be achieved on 2D displays. 
        \begin{figure}[h]
        \centering
       		\includegraphics[width=0.8\textwidth]{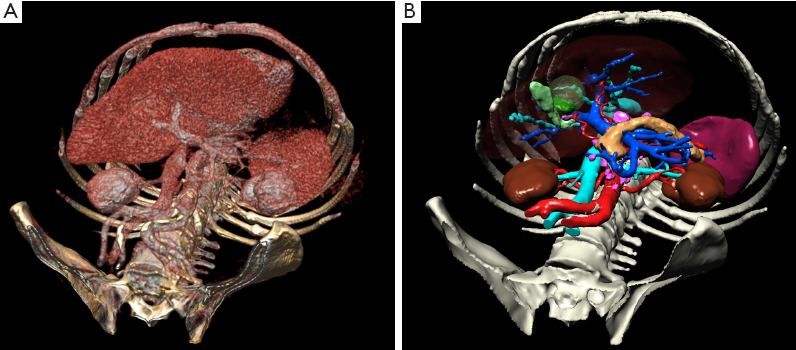}
			\caption[Comparison between volume rendering and surface rendering]{Comparison between volume rendering (A) and surface rendering (B) \cite{m32}.}
            \label{fig: Surface rendering}					
		\end{figure}
    \newline \indent
    \textbf{Volume rendering} \cite{volren1, volren2} helps visualize complex 3d images. Using a transfer function, different body structures with different pixel intensity values can be displayed with different color and intensity, for example, blood vessel voxels can receive one color and voxels corresponding to bone density can receive a different color. In the case of overlapping structures, this display method is limited when used on 2D displays \cite{2Dvolrenlim2, 2Dvolrenlim1}. \newline \indent
    A visual comparison between the volume rendering and surface rendering techniques is presented in Figure \ref{fig: Surface rendering}. \newline \indent
    The above stated limitation can be minimized using depth \textbf{3 dimensional (D3D)} imaging, i.e. displaying stereo 3D images on AR, MR or VR headsets. D3D transforms and displays cross sectional images on AR and VR headsets. This involves a  rendering engine heavily reliant on the GPU, which generates different left and right eye views to provide a true 3D visualization on the MR or VR HMDs. The rendering engine also provides some maneuverability possibilities to the user such as moving the viewing position, rotate or scale. The controller can be hardware such as an Xbox gamepad controller, or for example in the case of HoloLens, voice and hand gestures. \newline \indent
    \subsection{AR, VR and MR in medical imaging} \label{arvrintro}
    AR and VR provide enhanced viewing including depth perception and improved human machine interface (HMI) [12, 13]. AR, mixed reality (MR) and VR head mounted displays (HMDs) present a unique image for each eye, thus achieving  stereoscopy and depth perception. 
        \begin{figure}[h!]
    \begin{minipage}{\textwidth}
       		\includegraphics[width=1\textwidth]{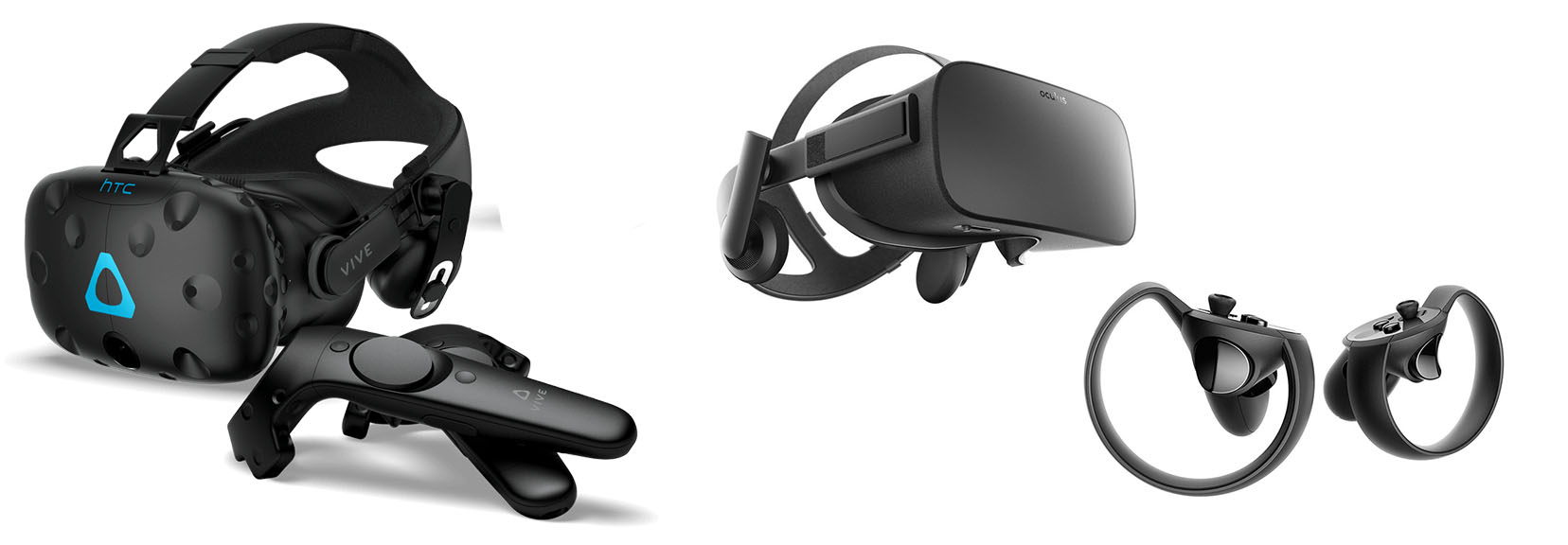}
			\caption[HTC Vive and Oculus Rift]{HTC Vive (left) and Oculus Rift (right). \footnote{Sources: \url{https://www.vive.com/eu/} (left); \url{https://www.oculus.com/} (right).}}
            \label{fig: oculusvive}
	\end{minipage}
    \end{figure}
     
    \textbf{Virtual reality} has been used to create an immersive and interactive environment. Virtual Reality (VR) can be full, semi, or non immersive [2]. Full immersive VR (ex. Oculus Rift and HTC Vive, illustrated in Figure \ref{fig: oculusvive}) displays a virtual image while the real surroundings are excluded from view [3]. 
   
\begin{figure}[h!]
    \begin{minipage}{\textwidth}
       		\includegraphics[width=1\textwidth]{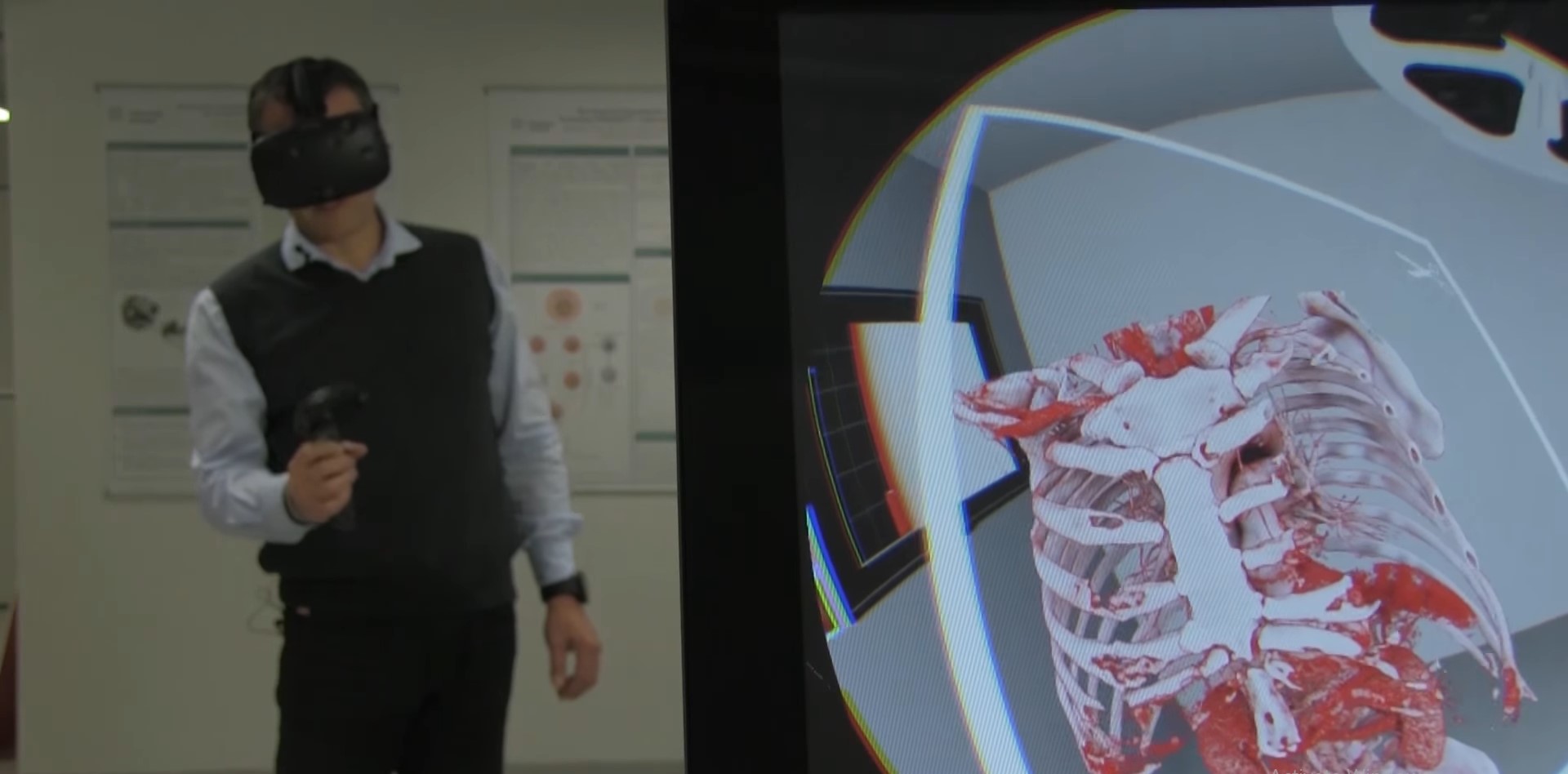}
			\caption[SpectoVive - University of Basel]{SpectoVive - University of Basel.\footnote{Source: \url{https://www.unibas.ch/en/News-Events/News/Uni-Research/Virtual-Reality-in-Medicine.html}}}
            \label{fig: baselvr}
	\end{minipage}
    \end{figure}

\begin{figure}[h]
    \begin{minipage}{\textwidth}
       		\includegraphics[width=1\textwidth]{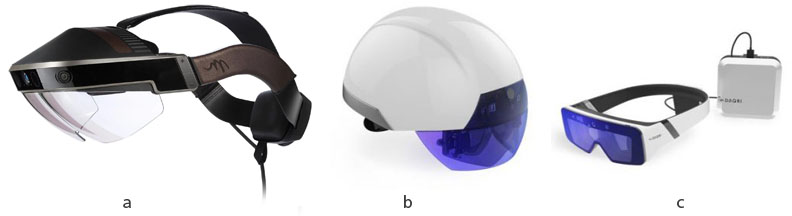}
			\caption[HoloLens' competitors: Meta, Daqri Helmet and Daqri Glasses]{HoloLens' competitors: Meta (a), Daqri Helmet (b) and Daqri Glasses (c).\footnote{Sources: \url{http://www.metavision.com/} (a); \url{https://daqri.com/blog/a-look-ahead-at-hannover-messe/} (b, c).}}
            \label{fig: metadaqri}
	\end{minipage}
    \end{figure}

\begin{figure}[H]
           		\center
		        \begin{minipage}{0.6\textwidth}
       				\includegraphics[width=1\textwidth]{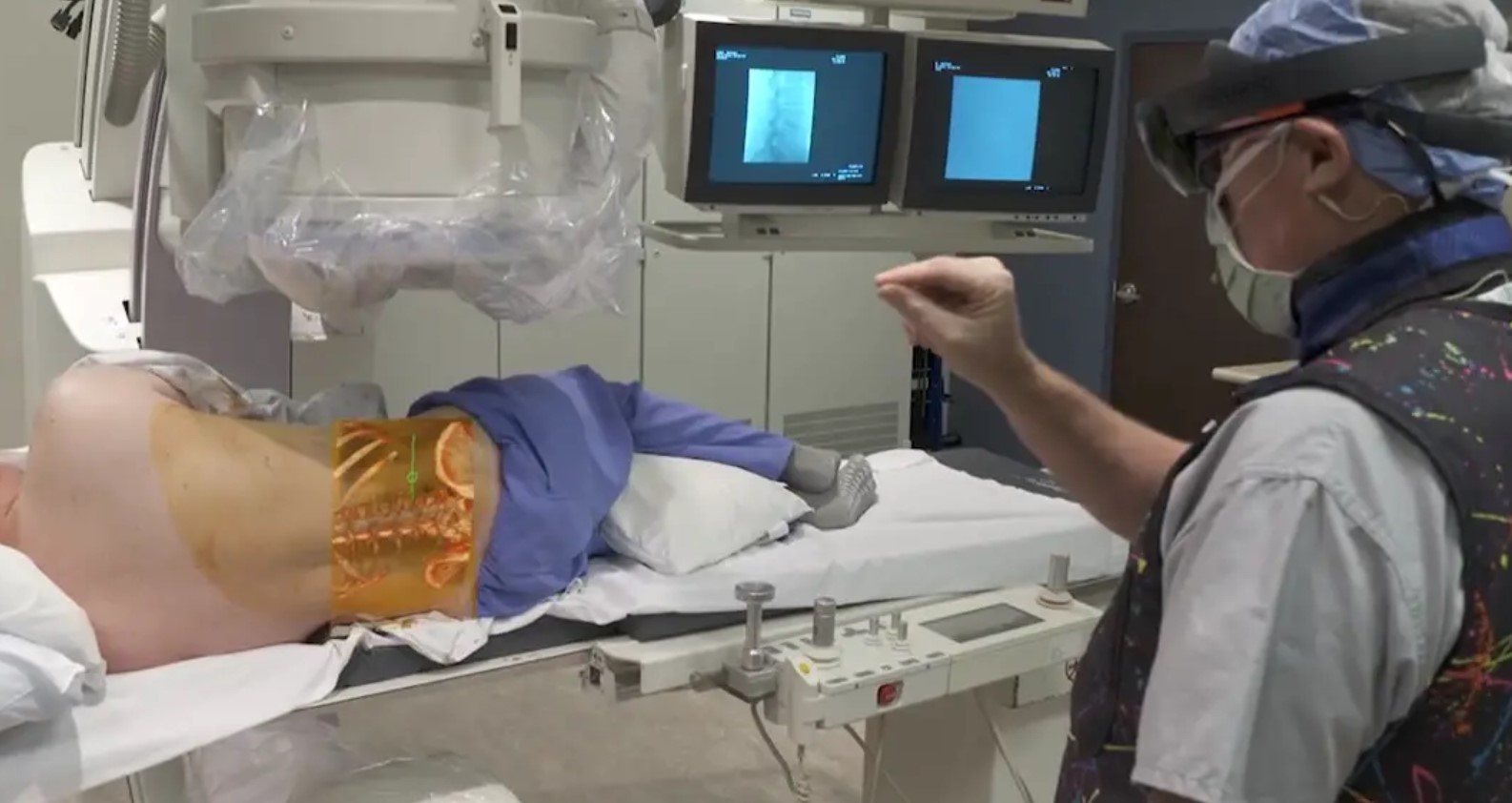}
					\caption[NOVARAD]{NOVARAD.\footnote{Source: \url{https://vimeo.com/223168345}}}
            		\label{fig: novarad}	
        		\end{minipage}
			\end{figure}

    Semi immersive VR (ex. Samsung Gear VR) displays the virtual image while the real world is partially occluded from view [3]. An example of non-immersive VR is a desktop computer. VR users can navigate through the virtual world by head movement, using the HMD tracking or by walking (using external camera tracking). Other ways of interaction with the virtual environment can be voice, gestures or hardware controllers. \newline \indent 
    Different visualization and interaction techniques have been developed: The University of Basel has developed the "SpectoVive" VR medical visualizations project, illustrated in Figure \ref{fig: baselvr}, which uses a hardware controller and places the user in a dedicated virtual room \cite {max5}, while in the US, the "DICOM VR" \cite{dicomvr} has been developed; however both projects have some limitations such as the hardware controller and full immersion (can not be used during live interventions for reasons of hygiene and surroundings awareness). \newline
    \indent The limitation of current software and VR systems in medicine comes mainly from the user's concern towards the surrounding physical world instead of the virtual world. AR overcomes these by providing a simple and immediate user interface (UI) on top of the electronically enhanced physical world. \newline \indent
    \textbf{Augmented reality} (AR) is divided into AR systems and \textbf{Mixed Reality} (MR) systems. In AR and MR, the user wears an HMD which simultaneously displays the virtual image and the real user environment \cite{Douglas2,glauner2017identifying}. In the case of medical applications, when pre-planning, intraoperative procedures, or a physical examination requiring medical imaging are needed, the real world image can be the actual patient’s body. \newline 
    \indent As the AR and MR domains are relatively new and the features specter is relatively continuous, we could not find a formal definition of where the AR stops and MR starts. Depending on how much we want to strengthen the definition requirements, the boundary can move between AR and MR. Generally, AR systems "augment" the reality by overlaying an image such as for example some information regarding address, name or height of some real world object. The overlayed image is not spatially anchored (changes position as we maneuver in the real world) and it can be two dimensional. In MR,  the real and virtual worlds are blended and mixed together in order to blur as much as possible the  boundaries between the real-word and the virtual world: the virtual image is generally three dimensional and carries properties of real objects (position, rotation, speed; seems affected by gravity; collides with and is visually occluded by real-world objects; the user can interact with the virtual object and change its position or visual appearance). Probably by strengthening this model's requirements, the authors of \cite{arvsvr} also argue that in AR, the virtual image is transparent like a hologram, while in MR it looks solid, a definition which makes Meta and DAQRI examples of AR systems, while Microsoft HoloLens would be an MR example. We think we can however safely state that at this moment Microsoft HoloLens carries the most MR features: objects are 3 dimensional, look solid, are anchored in space, can be interacted with by voice and gestures, can "collide" with the real world, can be visually occluded by objects in the real world, the experience can be shared, the user can freely move in the real world as there are no cables or other devices attached or required by the system, and more.\newline
    \indent Meta \cite{meta}, presented in Figure \ref{fig: metadaqri}(a), is a tethered HMD, i.e. it needs to be connected to a PC through a physical cable and is conceived to work in a stationary context which means the user can't walk around in the real world environment while wearing it. This limitation determined us to exclude it for this project, although it has a far larger FOV than HoloLens, and it can directly leverage the PC's computing power.     			 
    \newline
    \indent DAQRI models, illustrated in Figure \ref{fig: metadaqri}(b, c) were conceived mostly for an industrial environment usage and were eliminated for physical characteristics improper for our project (either helmet or using a cable) and because they provide only gazing as a control option (no hand gestures for example). It was also argued that the Daqri Helmet was presenting some tracking problems \cite{daqrev}. We are not aware if  collaborative synchronization has been implemented yet, or if the new 2018 Daqri glasses model fares better than HoloLens in regards to tracking performance, however, the Daqri glasses are unsuitable for our scenario as the headset makes use of an external mini-computer weared on the hip and connected through a cable. \newline
    \indent Through HoloLens, the user can view the virtual image and interact with the real world scene at the same time. HoloAnatomy is an interesting HoloLens medical application, which however displays only 3D Models (surface-rendered hollow objects, and not volumetric images), and does not offer the possibility for manipulation \cite{holoanatomy}. \newline
    	\indent One more notable example of using AR or VR in medicine is EchoPixel, which proposes an interesting holographic solution, although with the limitation that it can not be used during a live situation due to the visualization method and hardware controller used \cite{echopixel}. \newline
    \newline \indent One of the most advanced HoloLens systems for medical imaging has been built by NOVARAD, see Figure~\ref{fig: novarad}, in the USA. Their product renders patients’ medical images (CTs, MRIs) in 3D and presents them in an interactive manner, also giving medical personnel the possibility to interact with the imaging data as well as see it displayed in the anatomically correct location on the patient. \newline
	\indent The advantages of AR and VR over traditional medical imaging display methods are:
    \begin{itemize}[noitemsep]
        \item True depth perception, which considerably improves the diagnostician's interpretation \cite{m33};
    	\item  The 3D characteristic, which subsequently provides the possibility of adding novel user interface (UI) and tools \cite{Douglas2} like for example 3D cursors and markers, and an overall increased efficiency in processing high amounts of information (patient data);
    	\item Introduces the possibility of improved human control \cite{m33} interface capabilities like motion tracking (the scene camera view angle and position updates with head movement and rotation) \cite{Douglas2}, use of more ergonomic controllers inspired from gaming, or even voice and gestures which would provide a complete hygienic environment in the case of live intra-operative situations. We explore the voice and haptic interaction in surgical settings which would allow the medical images to be viewed and manipulated without contact by making use of the HoloLens' gesture recognition capability;
        \item In the case of MR (ex: HoloLens), the possibility of co-registration of the medical image to the real patient's image gives the impression of seeing into the patient, creating a live perception of the medical image inside the patient \cite{m33}. This allows better precision of surgeon's intervention, as seen in Figure \ref{fig: novarad}.
      \end{itemize}
\pagebreak
      \indent Also, there are still some challenges faced by AR and VR in the medical imaging area:
      \begin{itemize}[noitemsep]
      	\item The perception of structures overlapping in the image is reduced but not eliminated \cite{Douglas2};
      	\item Only light headsets stand a chance of being accepted into the operations room (HoloLens may still be considered somewhat bulky and heavy) \cite{Douglas2};
      	\item Motion sickness is still a potential problem which can hamper the medical personnel's capacity to best performing the medical act \cite{Douglas2};
        \item The HoloLens, being still an early product, has a limited field of view (FOV), which forces the user to turn his head (possibly away from a live surgical intervention) if the displayed image is outside FOV \cite{m33};
        \item  The HoloLens tinted visor that covers the display dampens the ambient light and  decreases the efficacy of other potential diagnostic monitors involved \cite{m33}.
      \end{itemize}     
      \indent \indent Despite the inherent challenges faced by any new technology, AR and VR will continue to develop to finally put their footprints into real life, including the medical industry. If augmented by the PC's computing power, we believe that at this time HoloLens offers the best mix of advantages and disadvantages for the medical visualizations usage. \newline
      \indent Summarizing, for our project's use-case, we prefer HoloLens over other devices for reasons like: surroundings awareness; good tracking (low latency and nosea); the multiple input methods offered (voice, gestures, gaze) while not requiring any input hardware device (the best answer to-date to the strict hygiene requirements in the operations rooms); no cables or other devices attached or needed, which could hamper the doctor's movements; relatively lightweight; the possibility of sharing (multiple users viewing in the same time the same content), and support from Microsoft. \newline 
      \indent We aim to combine the advantages of MR, in particular the Microsoft HoloLens HMD, with volumetric image segmentation and rendering, in order to advance this technology closer to the live operation rooms.
     
%$$$$$$$$$$$$$$$$$$$$$$$$$$$$$$$$$$$$$$$$$$$$$$
%$$$$$$$$$$$  CHAPTER 3 - MEDICAL IMAGE SEGMENTATION $$$$$$$$$$$$$$$$$$$$
%$$$$$$$$$$$$$$$$$$$$$$$$$$$$$$$$$$$$$$$$$$$$$$$
\chapter{Medical Image Segmentation}
	In regards to the medical image segmentation project requirement, we started with a liver segmentation method based on the \textit{K-means clustering algorithm}, and written in python. Because the results achieved were not satisfactory due to the algorithm's limitations (lack of context) towards performing almost real-time and without human intervention a medical image semantic segmentation of body organs such as the liver, we investigated a more complex approach, using ANN solutions.
    \section{The $K$-means clustering algorithm}
\begin{figure}[h!]
       	\includegraphics[width=1\textwidth]{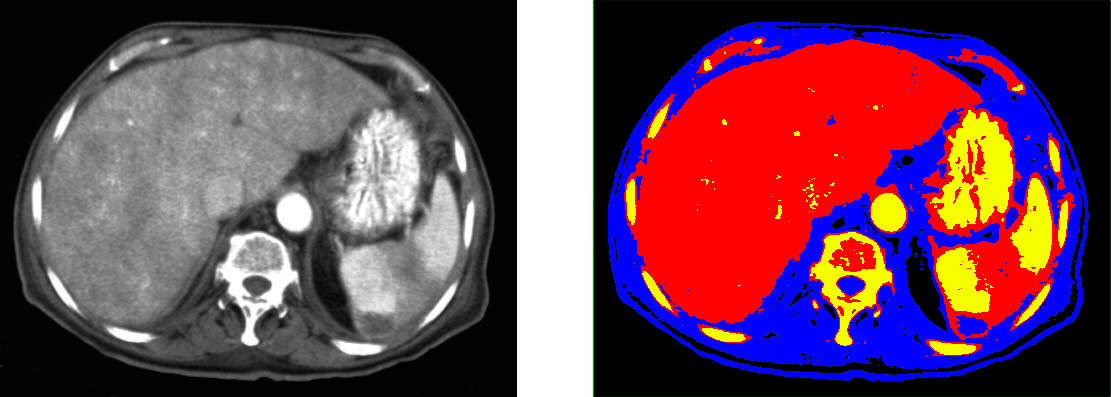}
			\caption[$K$-means: the initial image and the result for $K=4$]{$K$-means: the initial image (left) and the result for $K=4$ (right).}
        \label{fig: congA}
	\end{figure}

\begin{figure}[h]
     		\centering
       		\includegraphics[width=0.9\textwidth]{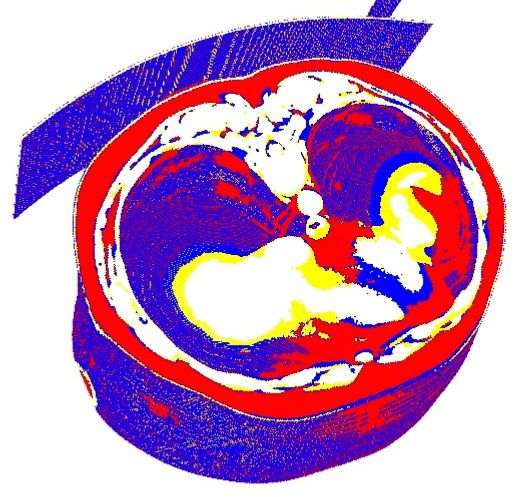}
			\caption[3D $K$-means: the result for $K=4$]{3D $K$-means: the result for $K=4$ (fourth color is white).}
            \label{fig: 3dkmeans}
	\end{figure}

    \indent Clustering is one of the most important unsupervised learning problems, which involves the grouping, or clustering, of data points. Given a set of data points, a clustering algorithm classifies each data point into a specific group.  A cluster is a group of objects which are “similar” between them and different from the objects from other clusters. In the given case, the similarity is measured in terms of intensity level similarity between the voxels in the image. The $K$-means algorithm doesn't involve any objective or loss function. \newline 
    \indent We have experimented with different values for the $"K"$ parameter: $K=3$ seemed to be too small as it could not separate a minimum of the required relevant features in the image, while starting with $K=5$, the result image started to be too fragmented feature-wise. The 2D result for $K=4$ is presented in Figure \ref{fig: congA}. We also tried this algorithm in 3D on a CT volume for $K=4$, as shown in Figure \ref{fig: 3dkmeans}. \newline 
    \indent As it can be seen in the Figures \ref{fig: congA} and \ref{fig: 3dkmeans}, because the algorithm uses the pixels' or voxel's luminance levels to decide to which class each pixel belongs, and because multiple anatomical structures can have similar HU levels, this algorithm can not be used as such for the task of liver segmentation without any human intervention. \newline
    \indent While some approaches to perform the liver segmentation using the $K$-means algorithm exist in the literature, they still require some minimal human intervention \cite{kmeans1} or excessive computational power \cite{kmeans2}, and consequently we had to explore more advanced solutions like a deep learning algorithm.
    %+++++++++++++++++++++++++++++++
    
    \section{The deep-learning algorithm}
\begin{figure}[h!]
       	\includegraphics[width=0.9\textwidth]{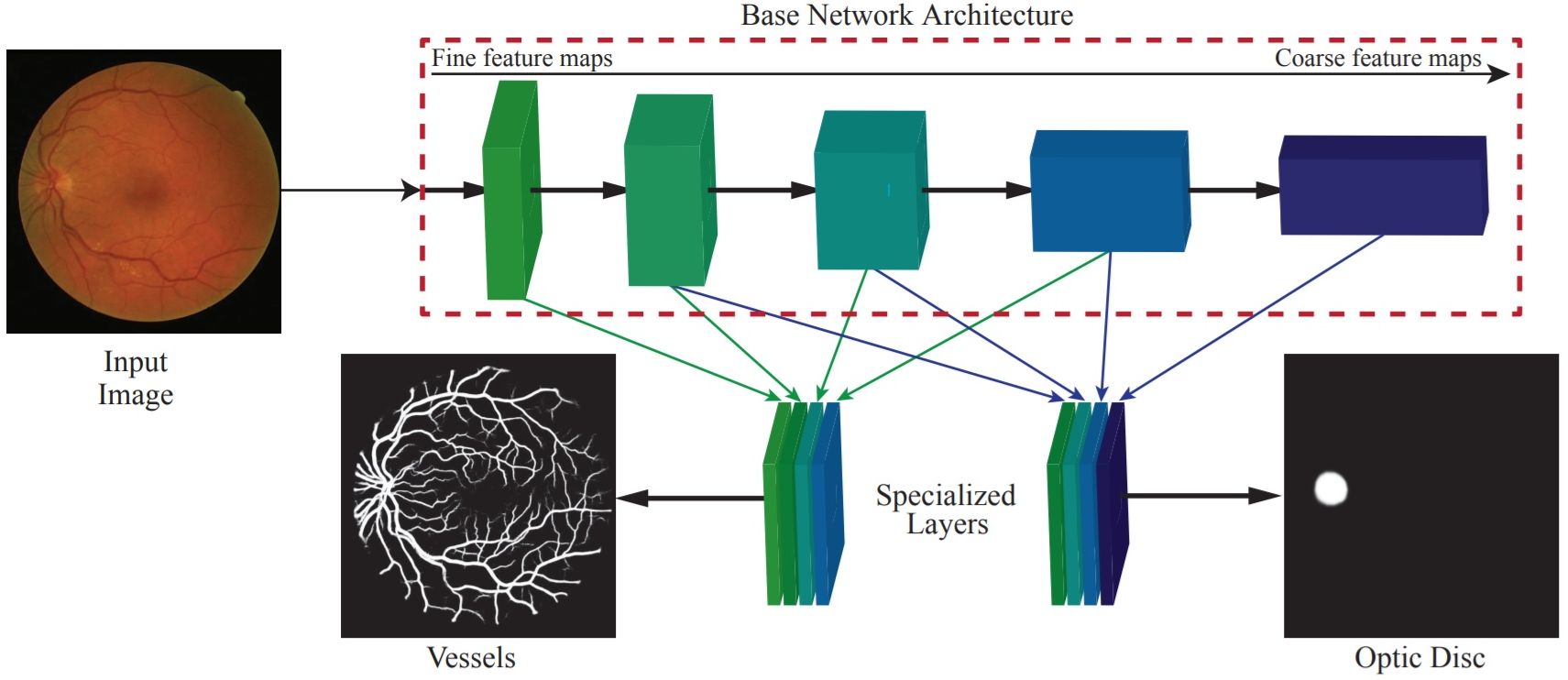}
			\caption[DRIU overview]{DRIU overview \cite{m24}.}
        \label{fig: DRIUarch}
	\end{figure}

	\indent The deep learning model used \cite{miriam} is based on the Deep Retinal Image Understanding (DRIU) network \cite{m24}, a model which performs the segmentation of the eye's blood vessels and optical disc. An overview of DRIU is presented in Figure \ref{fig: DRIUarch}.  After extracting side features, DRIU builds special
layers for the segmentation of arteries (left) and optical disc segmentation (right).
    Using convolutions, activation functions and max poolings, the architecture is itself inspired from VGG-16 \cite{m28}. As such, the base network is VGG-16 pretrained with Imagenet \cite{m7} and is made of convolutional stages which activate at same feature map resolution and are separated by pooling layers. At deeper levels, the information is more general and the learned features are closer to semantics \cite{miriam}. Being fine structures, the blood vessels benefit from the less deep layers, where the information is more granular. In the same time, the optic disc is advantaged by the deeper layers and more general features. Thus, DRIU uses several side outputs (convolutional layers connected at the end of a specific convolutional stage of the base network) specialized in different feature levels and the output is obtained by scaling and combining the side outputs.
         \newline
	\subsection{Deep learning algorithm architecture}
\begin{figure}[h!]
   			\begin{minipage}{\textwidth}
       		\includegraphics[width=1\textwidth]{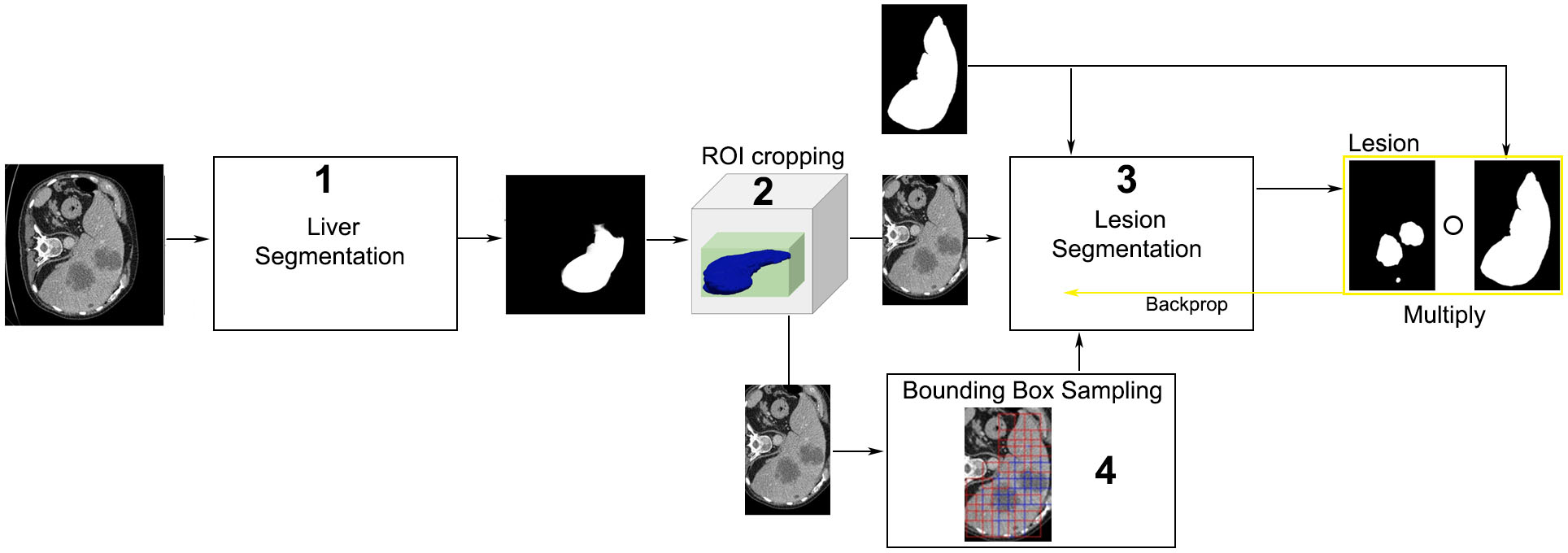}
			\caption[General algorithm overview]{General algorithm overview: after the liver segmentation (1), the segmented volume is cropped slice by slice around the liver ROI (2, green box). The resulting smaller liver segmentation volume is fed to both the lesion segmentation network (3) and the lesion detector (4). In the end, the output predicted by the lesion segmentation network is compared with the lesion detector's output, and only if both agree, the lesion localization is kept. Some elements in this figure are from \cite{miriam}.}
            \label{fig: GenArch}
        	\end{minipage}
		\end{figure}

\begin{figure}[h]
       		\includegraphics[width=1\textwidth]{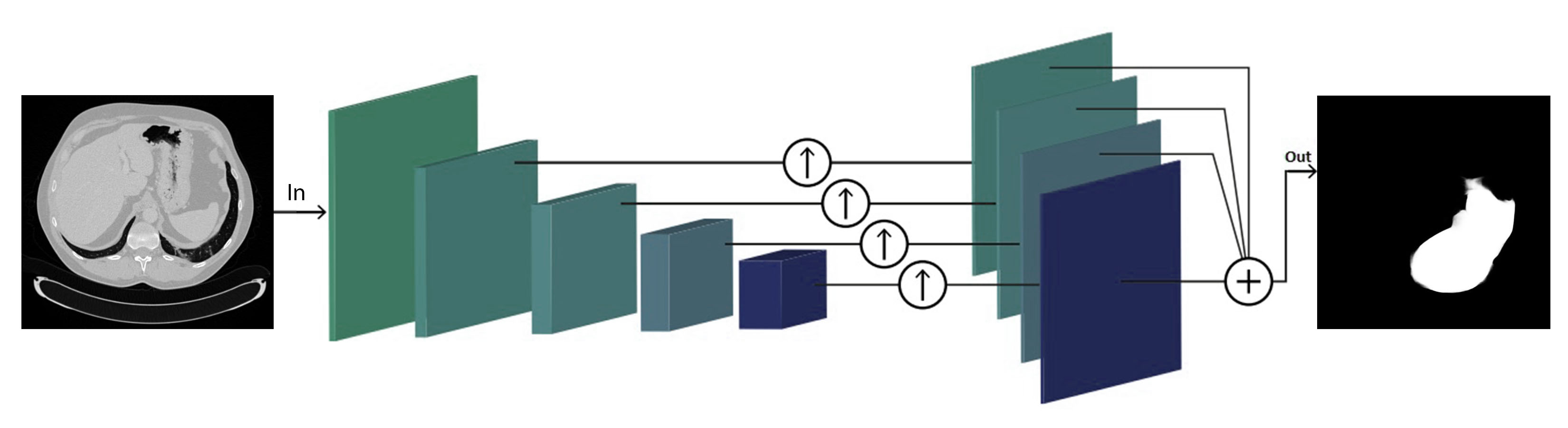}
			\caption[Liver segmentation architecture]{Liver segmentation architecture \cite{miriam}.}
            \label{fig: architecture}
	\end{figure}

    Being inspired from DRIU, the algorithm presented uses side outputs after each convolutional stage, all contributing to the final output. As the algorithm performs also lesion segmentation, the general architecture comprises multiple modules: liver segmentation, ROI cropping, lesion detector, lesion segmentation. An overview of the general architecture is presented in Figure \ref{fig: GenArch}, and a detail of the liver segmentation architecture is presented in Figure \ref{fig: architecture}. The lesion segmentation architecture starts from the same architecture used for liver segmentation, with different added enhancements.\newline
    \indent The most important aspects of the algorithm are:
   \begin{itemize}[noitemsep]
   		\item Pre-processing;
        \item Binary cross entropy loss weighting (for lesion segmentation);
        \item Using 3 consecutive slices as input;
        \item Post-processing (3D CRF).
   \end{itemize}
   
    \subsection {Pre-processing}
    The pixel intensities on the LiTS liver scans can have values exceeding |1000|. Many pixels with value -1024 belong to the background. Pre processing consists in clipping the pixels intensities values at min-max values that statistically belong to the liver and liver lesions (range -150 to 250). Afterwards, a min-max normalization is performed on each volume:
    $$z_i=\frac{x_i - min(x)}{max(x)-min(x)},$$
    which means we map the range between the old min and max values to the new range between the new min and max values (-150, 250).
	\subsection{Loss objective}
    The loss objective is defined using the Binary Cross Entropy (BCE) loss \cite{rothman2018artificial}: \newline
    \indent \[ L(y, \hat{y}) = - y \log \hat{y} - (1 - y) \log(1 - \hat{y}) ,\]
    \newline where \textit{y} is the actual truth and \textit{\^y} is the predicted value. The BCE provides a per pixel, individual loss, allowing to distinguish from losses coming from positive or negative truth pixels. As such, the positive and negative loss can be balanced separately. \newline 
    %\indent For the case when the algorithm is used for lesion detection, a BCE weighting has been introduced, in the following form:
    %\indent \[ L(y, \hat{y}) = - (1-w)y \log \hat{y} - w(1 - y) \log(1 - \hat{y}) ,\]
	\subsection{Input multiple 2D slices to take advantage of 3D data}
\begin{figure}[h]
       		\includegraphics[width=1\textwidth]{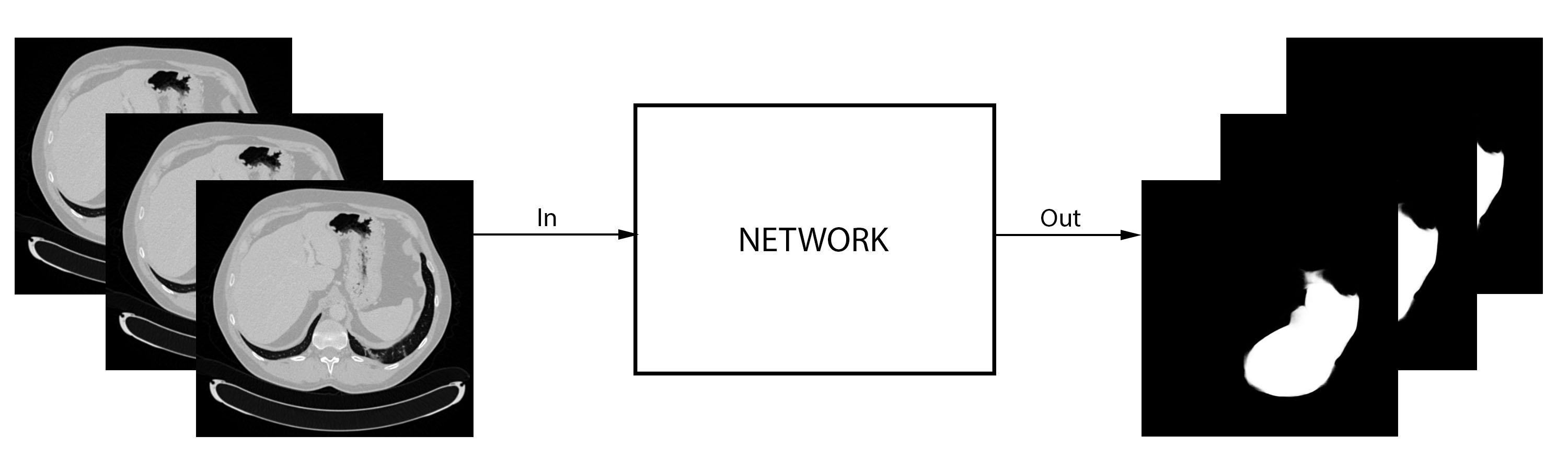}
			\caption[Input uses a series of three consecutive slices]{Input uses a series of three consecutive slices. For the test, only the middle output image is used.}
            \label{fig: 3slices}
	\end{figure}

    Originally the algorithm parsed the data as if the slices were independent, while in fact they have a strong 3D coherence. Also, the algorithm was pretrained with Imagenet, and it uses 3 channel images for training - RGB (Red, Green, Blue). Hence, as shown in Figure \ref{fig: 3slices}, the three channels can be fed simultaneously three slices of the volume analyzed, one at each RGB channel (2.5D approach), while during testing only the central output slice is kept. Figure \ref{fig: 3slices} refers to the liver segmentation, but the approach is the same for the lesion segmentation network.
\subsection{ROI cropping}
        After the liver segmentation, the outputted volume is fed to the ROI cropping, where the liver segmentation volume is cropped slice by slice around the liver ROI, as shown in Figure \ref{fig: GenArch}, module 2. The resulting smaller volume is afterwards used as input for the lesion detector and the lesion segmentation modules. 
        The number of positive pixels in each slice of the predicted liver masks resembles a Gaussian, so after a fitting of a Gaussian, a mean and variance are computed. The fitting is used to remove false positives, as all images outside a certain threshold are not likely to contain any lesion. As such, a significant number of false positives is removed, at the expense of a few false negatives.
%         \begin{figure}[h]
%         	\centering
%        		\includegraphics[width=0.5\textwidth]{ROIcropping.jpg}
% 			\caption[ROI cropping module]{ROI cropping module \cite{miriam}}
%             \label{fig: ROIcrop}
% 		\end{figure}
    \subsection {Using the liver segmentation for the lesion segmentation}
    	The liver segmentation is used in the process of lesion segmentation as a mask which limits back-propagation only to those pixels which belong to the liver segmented ROI. This way, only those pixels which can belong to the liver are used for the lesion segmentation learning process, and also, the process is more balanced as there are less negative pixels in the image. For this stage, the balancing term includes just the pixels contained inside the liver. This process is illustrated in Figure \ref{fig: GenArch} (module 3 - right). \newline
        \indent The possible disadvantage of this approach is that if the liver segmentation is not of good quality, this will negatively affect also the process of lesion segmentation.
        
    \subsubsection{Masking}
        As depicted in Figure \ref{fig: GenArch}, to perform the lesion segmentation, only those pixels predicted as belonging to the liver are considered. This improves the detection accuracy at the expense that if there is a mistake in the liver segmentation , this will propagate.
        
    \subsection{Lesion detector module}
     Because as such, the original algorithm was not able to consider a global view when performing lesion segmentation, false positives were triggered too often. In order to help the architecture to get a more general context, a lesion detector was added.
        \begin{figure}[h]
    	\centering
   		 \begin{minipage}{0.6\textwidth}
       		\includegraphics[width=1\textwidth]{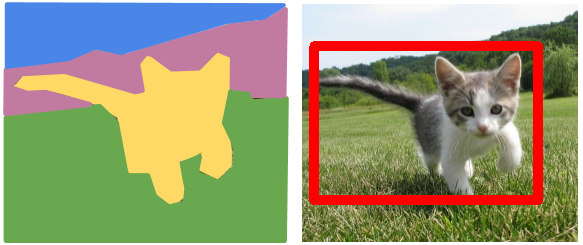}
			\caption[Segmentation vs detection]{Segmentation vs detection. \footnote {Source: \url {https://luozm.github.io/cv-tasks} } }
            \label{fig: segdet}		
   		 \end{minipage}
    \end{figure}
   
\begin{figure}[h!]
    	\centering
   		 \begin{minipage}{0.35\textwidth}
       		\includegraphics[width=1\textwidth]{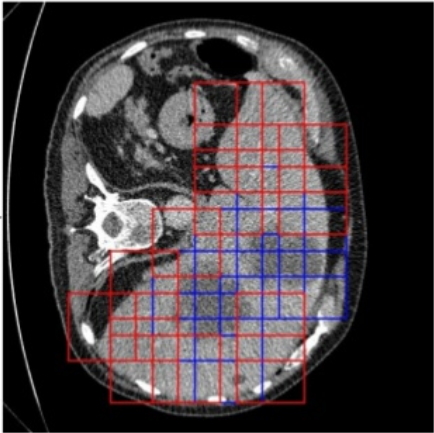} 	
   		 \end{minipage}
		\caption[The lesion detector]{The lesion detector \cite{miriam}. }
         \label{fig: detector}	
    \end{figure}

    \indent The difference between segmentation and detection is that image semantic segmentation classifies each pixel of the image as belonging to one class or another, while detection searches for some object (dog, car, lesion, smile) and localizes it generally with a bounding box. This is illustrated in Figure \ref{fig: segdet}.
    \newline \indent The detector was added in order to know, from a more global point of view, in which parts of the image there is actually a lesion. Then, the segmentation result is compared with the detector result, and only those locations where the results are in sync are kept. \newline \newline
    \indent The detector works as a sliding window, placing bounding boxes over the areas where it predicts a lesion. The window is 50x50 pixels, with a margin of 15 for more context, resulting in an overall size of 80x80 pixels. As illustrated in Figure \ref{fig: detector}, a box is placed if it overlaps with the liver on at least a 25\% area, and it is considered as positive if at least 50 pixels are considered as lesion pixels. 
    \newline \indent The lesion detector module is built from the pre-trained ResNet 50 model, without the Imagenet classification. Finally, only one neuron is used to take the healthy or non-healthy decision.
	
    \subsection{3D - Conditional random fields}
    A 3D Fully Connected Conditional Random Field (3D-CRF) is used for final processing. CRFs model the conditional distribution of the prediction by taking into account all the input. The 3D CRF is applied in order to refine the segmentation by taking into consideration the spatial coherence and also the input volume's pixels' intensities \cite{miriam}.
	The 3D CRF uses the algorithm from \cite{m6}.   \newline
    \indent The 3D CRF model uses as input the soft prediction of the network output and the pre-processed volume. The implementation is the one from \cite{m6}, and uses the mathematical representation proposed by \cite{m16}, which models a graph $G = (\nu,\nu)$ with vertices $i,j\in\nu$ representing each voxel in the image, and edges $e_{i,j}\in\varepsilon=\{(i,j),\text{ }\forall{i,j}\in\nu\textit{ s.t. }i<j\}$ between all the graph's vertices. \newline
	\indent The energy function is represented by Equation \ref{eq: crfenerg}, where $x$ represents each vertex's label. 
    \begin{equation} \label{eq: crfenerg}
    E(x)=\sum_{i\in\nu}\phi_i(x_i) + \sum_{(i,j)\in\varepsilon}\phi_{ij}(x_i,x_j)
    \end{equation}
    \indent Equation \ref{eq: crfunary} represents the unary potentials, where $I$ is the input image's intensity. 
    \begin{equation} \label{eq: crfunary}
    \phi_i(x_i)=-\log{P(x_i|I)}
    \end{equation}
    \indent Further, the potentials are computed pairwise as depicted in Equation \ref{eq: crfpot} \cite{m6}, where $\mu(x_i,x_j)=1,$ with $(x_i\neq{x_j}),$ is the Potts function; $p_i$, $p_j$ are the positions; $|p_i-p_j|$ the distance between voxels; $|I_i,I_j|$ is the intensity difference between the color vectors $I_i$ and $I_j$, and $w_{pos}$ and $w_{bil}$ represent linear combination weights. 
    \begin{equation} \label{eq: crfpot}
    	\phi_i(x_i,x_j)=\mu(x_i,x_j)\left(\underbrace{w_{pos}\exp\left(-\frac{|p_i-p_j|^2}{2\sigma^2_{pos}}\right)}_{\textit{smoothness kernel}}+\underbrace{w_{bil}\exp\left(-\frac{|p_i-p_j|^2}{2\sigma^2_{bil}}-\frac{|I_i-I_j|^2}{2\sigma^2_{int}}\right)}_{\textit{appearance kernel}}\right)
    \end{equation} \newline
    \indent The smoothness kernel removes small isolated regions, while the appearance kernel is inspired by the observation that nearby pixels with similar color are likely
to be in the same class. The degrees of nearness and similarity are controlled by the parameters $\sigma_{bil}$ and $\sigma_{int}$ \cite{m16}. \newline
    \indent Using the parameters, the pairwise terms and range can be tuned. Because the size of the lesions is generally much smaller than the liver, a smaller range is used here for the 3D CRF post-processing of the liver lesion. \newline \newline 

%$$$$$$$$$$$$$$$$$$$$$$$$$$$$$$$$$$$$$$$$$$$$$$$$
%$$$$$$$$$$$ EVALUATION $$$$$$$$$$$
%$$$$$$$$$$$$$$$$$$$$$$$$$$$$$$$$$$$$$$$$$$$$$$$$
\section{Evaluation}
	\subsection{Dataset: LiTS}
\begin{figure}[h]
       		\includegraphics[width=1\textwidth]{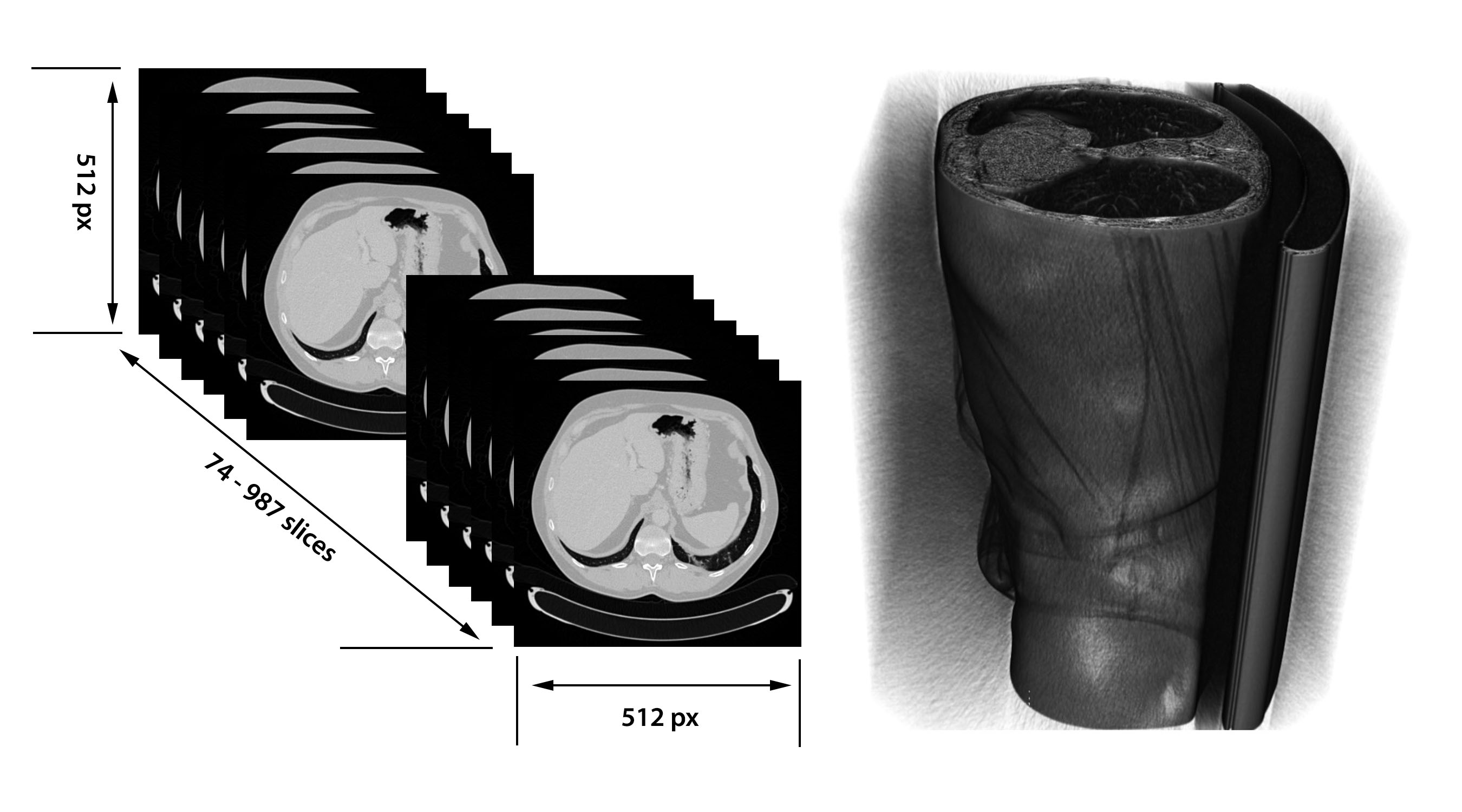}
			\caption[Structure of data volumes used, and actual 3D visualization]{Structure of data volumes used (left), and actual 3D visualization (right).}
            \label{fig: volumestruct}
	\end{figure}

    The liver dataset used to train the algorithm is obtained from the Liver Tumor Segmentation challenge (LiTS) \cite{LITS}, opened in 2017 for a ISBI workshop (International Symposium on Biomedical Imaging Conference). The challenge reopened once more during 2017, for a MICCAI workshop (Medical Image Computing and Computer Assisted Interventions Conference). The set consists of 131 CT scan volumes for training and 70 volumes for test. The file format is Neuroimaging Informatics Technology Initiative (Nifti), a format usually used for biomedical imaging. The file extension is ".nii", and it stores one volume per file. For encoding and decoding the file, a Matlab extension is used. Because each volume consists of multiple slices (74 up to 987 slices for each volume), the whole set includes in total 58,638 images. Each image is 512x512 pixels. The 3D image structure is depicted in Figure \ref{fig: volumestruct}.
    
   For the current task, the training set of 131 volumes was split into 80\% training and 20\% validation, i.e. 105 training volumes and 26 validation volumes.    
   %-----------------------------
	\subsection{Metrics}
	 The metric used to measure the results is the \textit{Dice (or the overlap index)}, which is the same as the \textit{F$_1$ score}. It is the harmonic average of the \textit{Precision} and \textit{Recall}, where an \textit{F$_1$ score} reaches its best value at 1 (perfect \textit{Precision} and \textit{Recall}) and worst at 0. 
     \indent \[ F_1= 2\times \frac{Precision \times Recall}{Precision+Recall}.\]
     $Precision$ is the number of correct positive results divided by the number of all positive results returned, and $Recall$ is the number of correct positive results divided by the number of relevant samples (all who should have been identified as positive). \newline \newline
     \indent The Dice score used is the $\overline{Dice}$, which is obtained by averaging the individual dice scores corresponding to each patient (volume), or the "dice per case". This is also one of the metrics assessed by the LiTS challenge. 
        %\item $Dice(\overline{P},\overline{R})$, calculated from averaging the \textit{Precisions} and \textit{Recalls} of each volume (sum of individual values divided by the number of volumes). %used to draw the precision and recall curves
     %----------------------------
	\subsection{Experiments on the LiTS dataset}
\begin{figure}[h]
       		\includegraphics[width=1\textwidth]{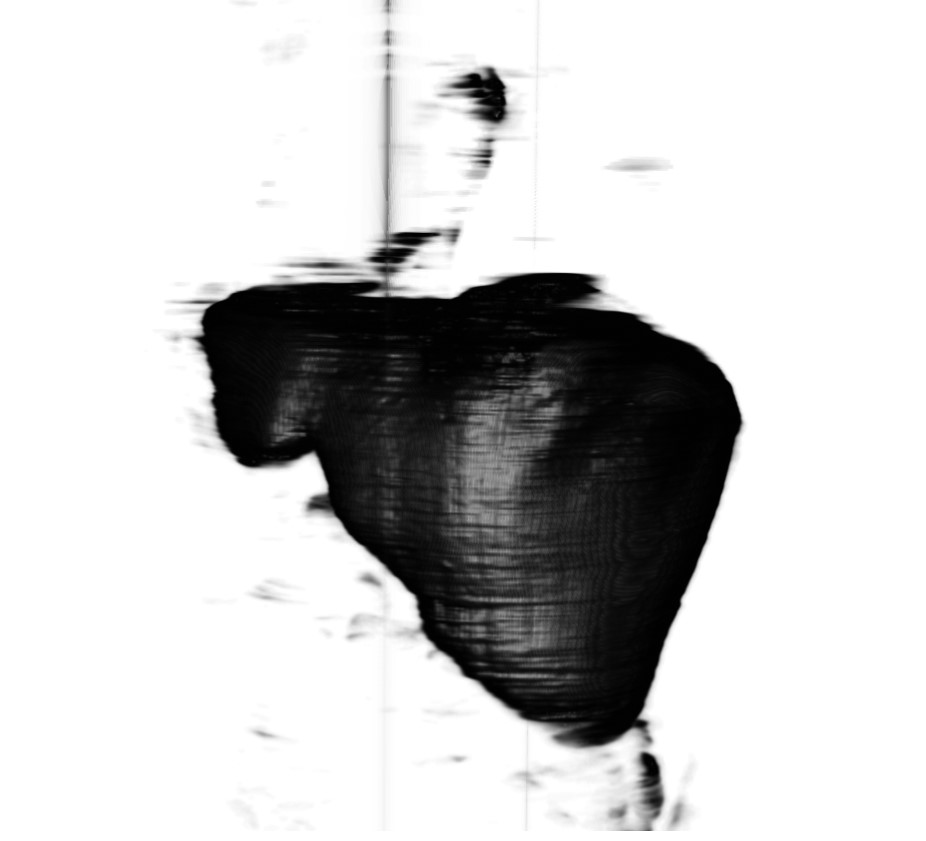}
			\caption[Sample liver segmentation result]{Sample liver segmentation result.}
            \label{fig: Assisting radiologists 2}
		\end{figure}

    	The training of the model was performed on a Linux PC and took approximately 6 days on a GTX950M GPU. On a Linux Desktop equipped with a GTX1080 GPU, training the liver segmentation network took around 30 hours. On a high-end GPU, we expect this to be accomplished in a matter of hours. \newline 
        \indent After training, the actual inference of a new segmentation is just a forward propagation through the network, which can take one or more minutes, depending on the GPU used. 
        A sample segmentation result is presented in Figure \ref{fig: Assisting radiologists 2}, where, for reasons of space, we cropped the image around the liver.
    
		\subsubsection{Loss Balancing}
       		\label{subsubsection:bce}
        	In our case, the data imbalance comes from the fact that not all liver scans contain lesions (most are healthy), and also from the fact that inside the liver, the healthy (negative) pixels are more than the positive pixels (lesion pixels). Imbalance of data falls into the general topic of biases in data sets, which is discussed in detail in \cite{glauner2017big,glauner2018reduction, glauner2018impact}. Because the negative pixels are the majority, the algorithm risks outputting all the image as negative, and hence, a new variable, $w$, is introduced in the BCE formula to compensate for this:
            \begin{equation}
            L(y, \hat{y}) = -(1-w)y\log\hat{y}-w(1-y)\log(1-\hat{y}).
            \end{equation}
            \indent This is implemented as a general balance factor, taking into account just the positive samples for each class. This is a global balancing factor and as such, all medical image volumes participate in the process of learning. Also, the different factors take into account only those images which contain the class. This is illustrated by the formulas ~\eqref{eq: generalbalancing} \cite{miriam}:
            \begin{equation}
				\label{eq: generalbalancing}
                \begin{gathered}
            w'_+=\frac{|\textit{Positive samples in V}|}{|\textit{Total samples in positive images of V}|} , \\
            w'_-=\frac{|\textit{Negative samples in V}|}{|\textit{Total samples in all images of V}|} , \\
            w_+=\frac{w'_+}{w'_-+w'_+} , \\
            w_-=\frac{w'_-}{w'_-+w'_+} .
                \end{gathered}
			\end{equation}
            
        \subsubsection{Results}
\begin{figure}[h]
    			\begin{minipage}{1\textwidth}
       				\includegraphics[width=1\textwidth]{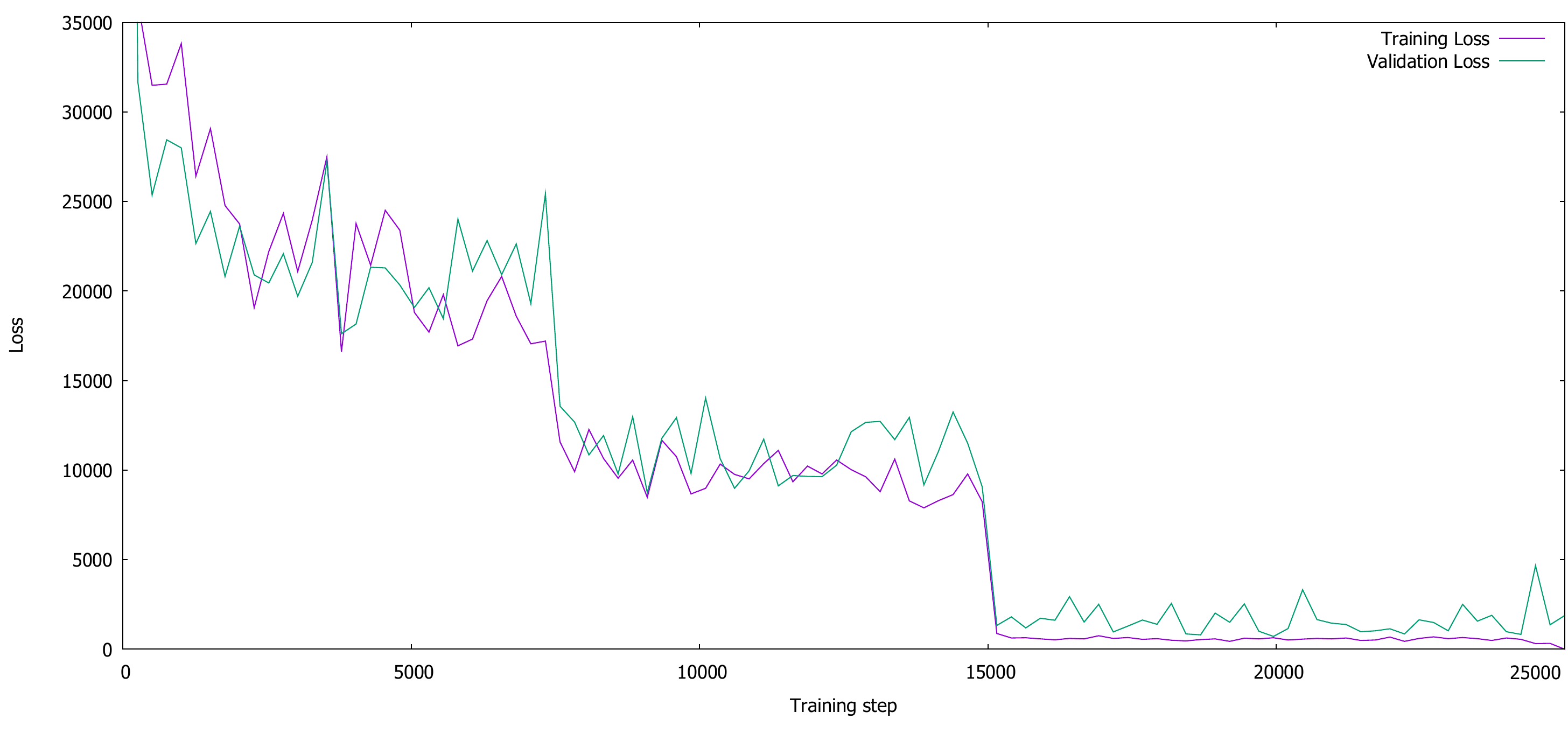}
					\caption[Liver segmentation: training and validation loss]{Liver segmentation: training and validation loss.}
            		\label{fig: trainloss}		
    			\end{minipage}
			\end{figure}

\begin{figure}[h]
            	\centering
       				\includegraphics[width=0.5\textwidth]{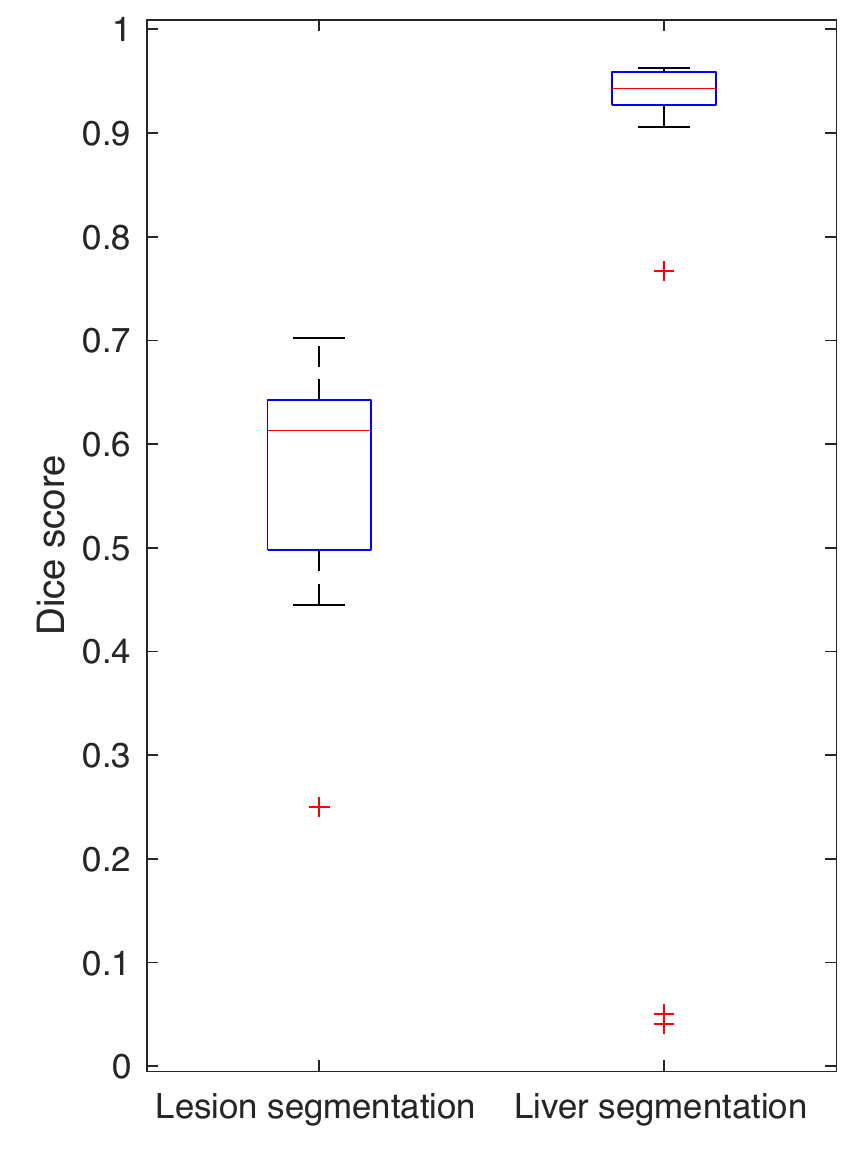}
					\caption[The LiTS challenge results comparison]{The LiTS challenge results comparison.}
            		\label{fig: lesliv}		
			\end{figure}

        The training evolution is somewhat noisy but the network steadily converges, as seen in the graph presented in Figure \ref{fig: trainloss}.
            \newline \indent For test on the LiTS challenge volumes, the lesion segmentation dice score is 0.586 while for the liver it reaches 0.938. \newline
            \indent For comparison, Figure \ref{fig: lesliv} and Table \ref{tab: lesliv} present the challenge results. The first position was occupied by the Lenovo Research team.
            \pagebreak
            \begin{table}[h]
            	\centering
				\caption[LiTS challenge: lesion and liver segmentation dice scores]{LiTS challenge: lesion and liver segmentation dice scores.}
				\label{tab: lesliv}
				\begin{tabular}{l|llllll}
Dice score\textbackslash{}Quantile & 0.025  & 0.25   & 0.5    & 0.75   & 0.975  & Best    \\
\hline
Lesion                      & 0.289 & 0.498 & 0.613 & 0.643 & 0.699 & 0.702 \\
Liver                       & 0.043 & 0.927 & 0.943 & 0.959 & 0.967 & 0.963
				\end{tabular}
			\end{table} 
            \newline \indent The segmentation times obtained during testing on the two GPUs available are presented in Table \ref{tab: testime}. Depending on the needs and budget, (after the initial training) a less powerful GPU can be also used for segmentation.
            \begin{table}[h]
            	\centering
				\caption[Training and testing times for the liver segmentation]{The training time for the liver segmentation algorithm, and the average test time per volume for the two GPUs assessed.}
				\label{tab: testime}
				\begin{tabular}{l|ll}
				GPU\textbackslash{}time & training & testing \\
                \hline
				GTX950M & 135 hours & 159 sec \\
				GTX1080 & 30 hours & 35 sec
				\end{tabular}
				\end{table}
        \subsubsection{Data preparation for visualization} For each volumetric image, the output liver segmentation is a folder containing a set of image slices. At the moment we manually convert the slices to the ".raw" 3D format for loading and visualization in the Unity game engine using "ImageJ", an external application (rendering on the Desktop PC Server and visualization on HoloLens).
        \section{Discussion}
        In this chapter, we have provided and evaluated an algorithm for segmenting the liver. Our experimentation validates that using the 2.5D algorithm with pre-trained weights is feasible for our project goal of a near-real-time automatic liver segmentation. \newline
        \indent The strategies used by the algorithm which contribute to improving the final segmentation results, are: the balancing strategy, the 2.5D approach, volume pre-processing, backpropagation only through the liver for lesion segmentation, applying the detector and the 3D-CRF as a final post-processing step. The method of limiting the learning process only to those pixels belonging to the liver is somewhat similar to the "attention" strategy, as only a specific region has been selected to focus the learning on.\newline
    \indent The segmentation network also has the potential to perform the segmentation of other different structures. As the algorithm is derived from DRIU which was able to perform also the blood vessels' segmentation, in the future it would be interesting to investigate how this can be adapted for the segmentation of other liver structures such as the arteries. \newline
    \indent While we are used to encountering in our machine learning tasks much larger datasets, the datasets specific to this task are usually much smaller. For example, the original authors \cite{miriam} have successfully trained and tested this algorithm also on the "Visual Concept Extraction Challenge in Radiology" (Visceral) dataset \cite{visceral} which comprises only 20 volumes, out of which 18 were used for training and 2 for validation. The reasons for such small datasets are related to the difficulty of producing these labeled volumetric images datasets, data privacy and data size.
 %$$$$$$$$$$$$$$$$$$$$$$$$$$$$$$$$$$$$$$$$$$$$$$
%$$$$$$$$$$$  CHAPTER 5 HOLOLENS VISUALISATION $$$$$$$$$$$$$$$$$$$$$$$$$
%$$$$$$$$$$$$$$$$$$$$$$$$$$$$$$$$$$$$$$$$$$$$$$$
\chapter{HoloLens Visualization}
    In this chapter, we will present our solution for the visualization of the 3D CT and MRI medical images using the Microsoft HoloLens MR HMD. We prefer MR over VR because it "keeps" the user (medical staff) in the real world, thus further limiting possible nausea (most of the image is the real world which has zero latency) and allowing the solution to be used in a dynamic situation. Moreover, we prefer HoloLens over other MR devices for practical reasons like the multiple user input possibilities offered, integrated Wifi connection (no additional cables or connected devices needed), lightweight (compared to other options) and for technical reasons like features and support. The GPU computing power limitation is addressed by using a dedicated Desktop Windows PC rendering server.
	\section[Hololens with external rendering]{Using Unity and WebRTC to deliver desktop rendering power to HoloLens}
    \begin{wrapfigure}{r}{0.4\textwidth}
    \begin{minipage}{0.4\textwidth}
       		\includegraphics[width=1\textwidth]{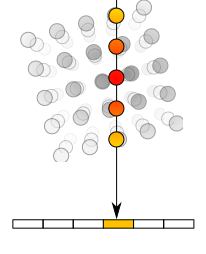}
			\caption[Volume raycasting]{Volume raycasting.\footnote{Source: \url{https://en.wikipedia.org/wiki/Volume_ray_casting}}}
            \label{fig: raycasting}		
    \end{minipage}
	\end{wrapfigure}
\begin{figure}[h]
       		\includegraphics[width=1\textwidth]{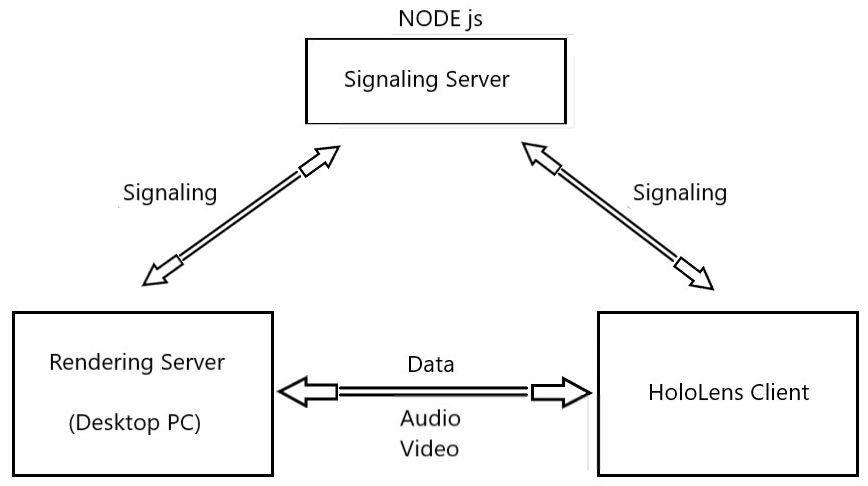}
			\caption[System logical architecture]{System logical architecture.}
            \label{fig: webrtc architecture}
	\end{figure}

    Being made out of voxels, medical imaging data volumes can be rendered by using a technique called raycasting. In Unity, a simple cube geometry is assigned a material which in its turn loads a shader. A shader is, in fact, a program with the difference that it is written for and it runs on the GPU. The shader loads the texture, in our case the 3D volumetric image made out of voxels, and tells the GPU how to render the object, in this case using the technique often called volumetric ray casting, ray tracing or ray marching. Basically, for each pixel of the final image, a "ray" is sent through the volume and the values of the nearby pixels intersected by the ray are interpolated to compute the final image pixel, as depicted in Figure \ref{fig: raycasting}. A usual medical volume size of 512x512x1024 voxels, i.e. more than 250 million voxels, is hence computationally intensive in terms of GPU performance and we needed to externalize these computations to a dedicated Windows server machine. \newline 
    \indent As a result, our visualization architecture comprises three interconnected applications, running in the same time: the Hololens Client, the Windows Desktop Server, and a Signaling Server which manages the communication and connection between the first two, as depicted in Figure \ref{fig: webrtc architecture}.

\begin{figure}[h]
       		\includegraphics[width=1\textwidth]{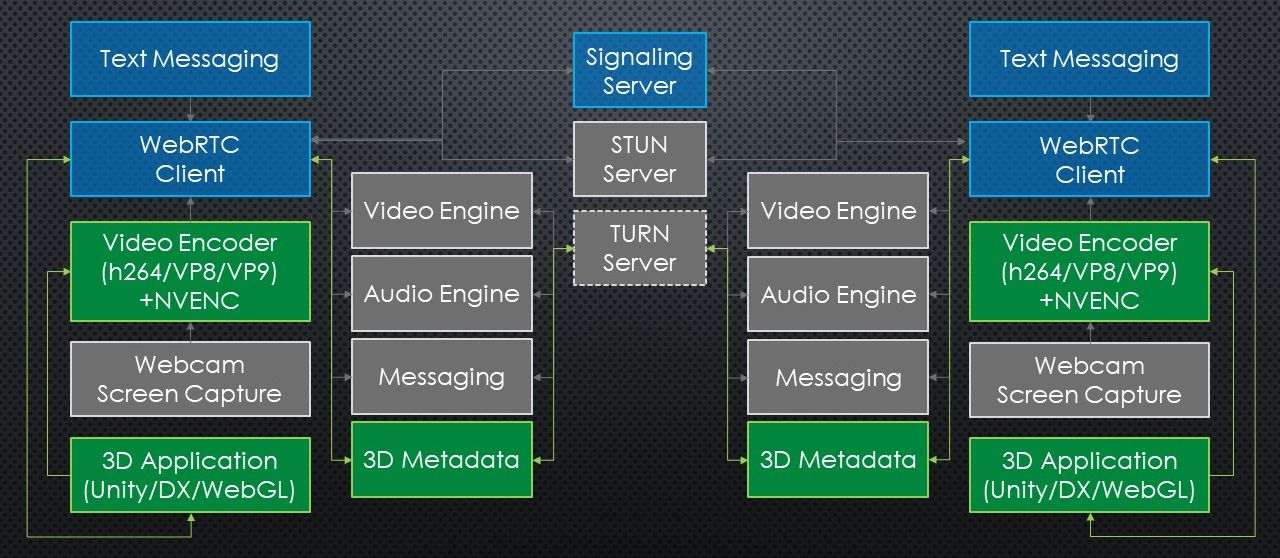}
			\caption[WebRTC extended by 3DToolkit]{WebRTC extended by 3DToolkit (green) \cite{catalyst}.}
            \label{fig: arch3dT}
	\end{figure}

    Our solution makes use of the "3D Toolkit" which uses the WebRTC (Web Real-Time Communications) protocols and API as well as the NVEncode hardware encoding library from NVIDIA. The system architecture is depicted in Figure \ref{fig: arch3dT}.
    
\begin{figure}[h]
       		\includegraphics[width=1\textwidth]{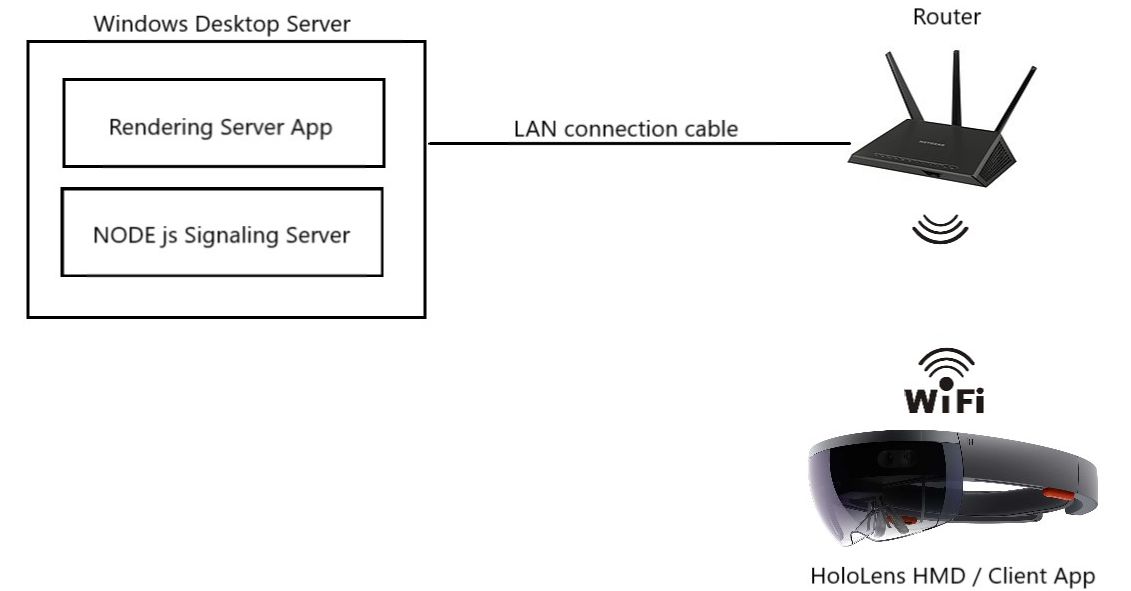}
			\caption[Hardware architecture]{Hardware architecture.}
            \label{fig: hardact}
	\end{figure}

    \indent The 3D Streaming Toolkit provides server-side libraries for remotely rendering 3D scenes, client-side libraries for receiving streamed 3D scenes, low-latency audio and video streams using WebRTC, as well as high-performance video encoding and decoding using NVEncode \cite{catalyst}. \newline
    \indent Among others, some of the necessary prerequisites are Windows 10 Anniversary Update, Visual Studio 2017, Windows 10 SDK - 10.0.14393.795, an NVIDIA GPU with NVIDIA drivers and CUDA Toolkit 9.1 (required for NVEncode) installed and Unity 2017.4.4f1 LTS release.  \newline
	\indent Concerning the hardware architecture, it comprises 3 components: a Router, the Desktop Windows Server (hosting the Rendering Server app and the Signaling Server app), and the HololeLens HMD running the DirectX HoloLens Client. The hardware architecture is presented in Figure \ref{fig: hardact}.
    
        The features provided are presented in Table \ref{tbl: features}:
        \begin{table}[h]
        	\centering
        	\small
            \caption[Client and server feature matrices]{Client and server feature matrices.}
			\begin{tabular}{l|lll}

Feature \textbackslash Platform    & HoloLens DirectX Client       & HoloLens Unity Client    & Unity Win32 Server       \\
\hline
AV streaming   			           & Y                        & Y                        & Y                        \\
Data streaming                     & Y                        & Y                        & Y                        \\
HTTP signaling                    & Y                        & Y                        & Y                        \\        
HTTPS signaling                   & Y                        & N                        & N                        \\
Signaling heartbeat               & Y                        & N                        & N                        \\
OAuth24D authentication             & Y                        & N                        & N                        \\

		\end{tabular}
		\label{tbl: features}
	\end{table}

    	\subsection{Signaling server and networking}
   		 For reasons such as control, reliability, transmission speed and latency we have preferred to use for communication the local network instead of the internet. \newline
         \indent The peers interact with the signaling server to share the handshakes and start a direct peer-to-peer transmission. After this point, the actual data is sent directly between client and server. While the traffic and computation load of the signaling server is low, it is still a core component of the WebRTC connection architecture. To simplify the overall architecture and improve communication speed, we have deployed the signaling server on the same windows desktop machine that runs the Rendering Server.
         \newline \indent The signaling server code is \textit{node.js} and is started with the simple command \textit{"node ./server.js"}.

		\subsection{Client}
        This is a simple DirectX client which connects to the Signaling Server for handshaking, to finally establish a peer-to-peer connection with the Rendering Server via WiFi in order to receive the rendered frames as a stream, and send back to the Rendering Server updates concerning the HMD's position and rotation via the dedicated data channel. The Rendering Server updates the view per the newly received coordinates of the HoloLens HMD in the world. 
        
        \subsection{Server}
        The server is built using the Unity game engine and is meant to offload the heavy GPU rendering task from the HoloLens client. It is meant to run in a Windows OS, and makes use of the following technologies: 
        
        \begin{itemize}[noitemsep]
        \item NVIDIA drivers and CUDA library to render and encode the scene frames which will be sent to the HoloLens client. Most NVIDIA graphics cards include dedicated hardware for video encoding, and NVIDIA's NVEncode library provides complete offloading of video encoding without impacting the 3D rendering performance. 
        \item The WebRTC open source project, released by Google in 2011 for the development of real-time communications between apps, including low latency VOIP audio and video applications. Communication between peers is managed through one or more data channels. The Video Engine is offered as a middleware service to establish a video data channel and to automate buffer, jitter, and latency management. The Audio Engine does the same in regards to audio transmission and is conceived for efficient processing of voice data. Applications can open different data channels for custom  messages \cite{catalyst}.
        \end{itemize}
        \indent  The 3DStreamingToolkit's additions to the typical WebRTC usage are:
        \begin{itemize}[noitemsep]
        	\item The NVIDIA NVEncode hardware encoder library for real-time encoding of 3D rendered content was added to the video encoders. 
       		 \item  A dedicated data channel manages the camera transforms and the user interaction events. This channel is used to update the HoloLens camera position in the rendering server when the user moves through the room.
        \end{itemize}
        \indent \indent These were implemented by means of plugins that engage Unity or native DirectX rendering engines. The Unity server makes use of a native plugin produced by the 3D Toolkit build pipeline. The plugin negotiates with clients to configure a stream, and for encoding and sending visual frame data from Unity to the client. \newline
        \indent The \textbf{core scripts} employed by the server are the \textit{StreamingUnityServerPlugin}, which provides a wrapper around the native plugin that powers the experience. An instance of the wrapper is created by the \textit{WebRTCServer}, and exposed publicly. The \textit{WebRTCServer} is the main WebRTC component, which configures  the native plugin and handles client input data. Finally, the \textit{WebRTCServerDebug} enables detailed logging data on request.
        
                \section{User interface}
         \begin{figure}[h]
                \centering
       		\includegraphics[width=0.8\textwidth]{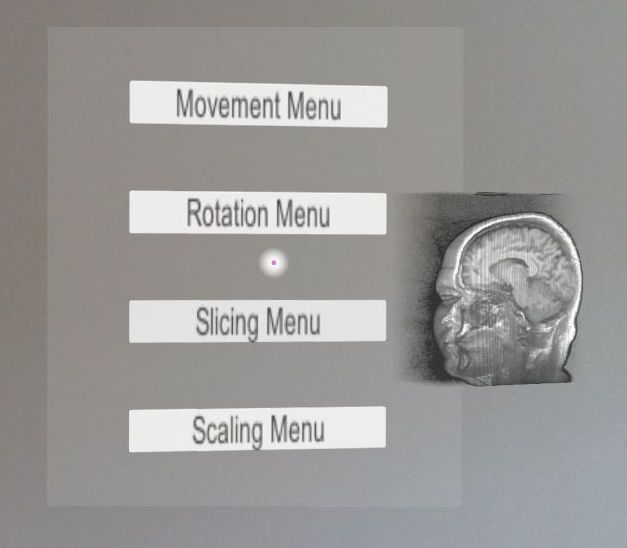}
			\caption[Application screenshot]{Application screenshot.}
            \label{fig: Assisting radiologists 1}
		\end{figure}
        At this moment we have a basic yet functional UI which allows movement, rotation and scaling of the 3D medical volumetric image visualized. These scripts are acting directly on the data cube. \newline
        \indent Another set of scripts which activates on the shader level allows volume slicing and image luminosity adjustments. A sample visualization is presented in Figure \ref{fig: Assisting radiologists 1}.
        
        \section{Performance}
        \indent The server executable is run through a .bat file, using the following command: \newline \newline
       \indent \textit{"server.exe -force-d3d11-no-singlethreaded"} \newline \newline
        \indent Tests started on a system equipped with a GeForce GTX950M NVIDIA GPU and an Intel i5 CPU. However, because the raycast drawing technique combined with the large data volume was too resource intensive for the given machine, we had to try more powerful solutions:
        \begin{itemize}[noitemsep]
       		 \item An external GPU enclosure, the "Akitio" node equipped with an NVIDIA GTX1070, proved to be too limited in connectivity options as it requires a machine equipped with a Thunderbolt 3 slot (typically Apple machines), or a very specific Windows machine model (generally laptops) with a USB 3.1 port, and as such it can not offer a scalable and flexible option.
       	 \item Running the executable on a cloud Windows machine was eliminated as it introduces more latency possibilities (important for real-time MR visualization) and uncontrolled layers in the process, such as an external uncontrollable windows cloud platform, and also the internet as an additional communication channel which is undesirable for our use case (a safe and reliable medical application), due to the unpredictable nature of the internet.
        \end{itemize}
        \indent \indent A dedicated local Windows desktop server equipped with an NVIDIA GTX 1080 GPU and an AMD octa-core CPU from the FX 8000 family running at 3.5Ghz, communicating over local WiFi proved to be the solution offering the best mix of control, reliability, scalability and performance. \newline
        \indent A comparison between the frame rates obtained by rendering a volume of 512x512x986 voxels, i.e. 260 million voxels, using the raycast volume rendering technique is presented in Table \ref{tab: GPU performance}.
          \begin{table}[h]
          	\centering
            \caption[Different GPUs' performance for the reference volume]{Different GPUs' performance for the reference volume.}
            \begin{tabular}{l|l}
            GPU          & FPS \\
            \hline
            HoloLens GPU & 1   \\
            GTX 950M     & 4   \\
            GTX 1080     & 18 
            \end{tabular}
            \label{tab: GPU performance}
          \end{table}

\section{Discussion}
	\indent We must notice that, as the experience involves stereoscopy, the scene is rendered twice - once for each eye. The 18fps obtained on the GTX 1080 GPU are stereo fps, hence the GPU has rendered in fact 36 mono fps. As such, we believe that this is a pertinent, scalable solution. Using 2 GTX 1080 in SLI mode should almost double the fps to around 35 stereo fps. Using superior graphics cards should also proportionally improve performance.

        \begin{figure}[h]%{0.8\textwidth}
        	\centering
    		\begin{minipage}{0.8\textwidth}
       			\includegraphics[width=1\textwidth]{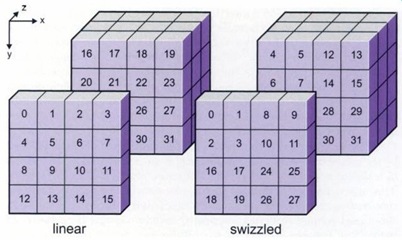}
           		 \caption[Improving GPU memory access]{Improving memory and cache efficiency. \footnote{Source: \url{http://graphicsrunner.blogspot.com/2009/02/volume-rendering-201-optimizations.html}}}
           		 \label{fig: GPUcache}		
   			 \end{minipage}
		\end{figure}

    \indent Until now we used brute force for rendering, so adding shader quality improvements like:
        \begin{itemize}[noitemsep]
        	\item GPU cache efficiency and memory access, as depicted in Figure \ref{fig: GPUcache}: as the volume data is linearly loaded into memory, a ray that is cast through the volume has poor chances to have the fastest possible memory access to neighboring voxels information as it traverses the volume. This can be improved by converting the layout to a block based layout. Thus, as we travel through the volume, we will be more likely to faster access in memory the neighboring voxels.
        	\item Eliminating the 0 alpha voxels: right now, we are ray-casting through the whole volume in the scene, i.e all voxels, even if the volume contains areas with zero alpha voxels. The zero alpha voxels can be skipped and we can render only the volume parts which contain non-zero alpha voxels.
       \end{itemize} 
   
    In conclusion, with the right hardware and some shader improvements, we believe that this solution can provide a stereo medical visualization at around 60 fps.
    Although not experimented during this project due to hardware constraints, the 3D Toolkit authors state that an unlimited number of peers can connect to a single instance of a server. However, this number will eventually be limited by the hardware, as NVIDIA enforces a maximum of 2 GPU encoding sessions on desktop series graphics cards. The number of peers can be limited via \textit{signaling}, more specifically, through configuring a ".json" file.
%$$$$$$$$$$$$$$$$$$$$$$$$$$$$$$$$$$$$$$$$$$$$$$
%$$$$$$$$$$$ $$$$$$$$$$$$$$$$$$$$$$$$$$$$$$$$$$$$$$$ FUTURE WORK $$$$$$$$$$$$$$$$$$$$$
%$$$$$$$$$$$$$$$$$$$$$$$$$$$$$$$$$$$$$$$$$$$$$$
%\chapter{Discussion}
%    \section{ML segmentation}
%%    OLD

%$$$$$$$$$$$$$$$$$$$$$$$$$$$$$$$$$$$$$$$$$$$$$$
%$$$$$$$$$$$  DISCUSSION $$$$$$$$$$$$$$$$$$$$$$
%$$$$$$$$$$$$$$$$$$$$$$$$$$$$$$$$$$$$$$$$$$$$$$
\chapter{Conclusion and Future Work}
In this project, we proposed a solution for building a software suite for automated machine learning segmentation of medical radiology images and advanced true-3D visualization of these images using the Microsoft HoloLens mixed reality head-mounted display. \newline \newline
	\indent At present, we have implemented and evaluated the project's main functionalities and algorithms such as: converting and loading the medical image file formats into the Unity 3D game engine; CT and MRI volumetric image visualization on HoloLens; a basic yet functional UI for image manipulation; offloading the heavy GPU rendering of volumetric images from the HoloLens client application to a dedicated Windows desktop server through local WiFi, and a fully automatic solution for the medical image segmentation using a deep learning algorithm. For training and testing of the algorithm, a labeled set of 200 liver scans has been used.  We have demonstrated the performance improvement and the possibility of scaling the HoloLens Client - Unity Desktop Server visualization solution using WebRTC and high-end GPUs. \newline \newline
    \indent Having studied the possibilities and the feasibility of using these solutions for practical medical visualization purposes, we are confident that the results are promising. \newline
            \indent While there is yet room for improvement, we are confident that this system has high potential for improving the quality, speed and efficiency of the medical act while lowering the overall operative costs, and also step up into the operations rooms in the near future. \newline \newline
            \indent Future work comprises performance improvements in regards to latency and frame, achievable mainly by scaling the hardware and secondly by improving the drawing technique (shader). While the key feature of automatic liver segmentation has been achieved, at the moment some minor steps are still manually performed (transferring, converting and loading data), so our future work will further focus on fully automating the pipeline so that little to no human intervention is needed. 
%---APPENDICES---------------------------------------------------
%\begin{appendices}
%\chapter{Visualisations}
%The contents...
%\end{appendices}
%================================================================
%================================================================
\backmatter
%================================================================
%================================================================
%-------------------------------------------------------------
% Bibliography

\bibliography{thesis.bib}
\bibliographystyle{ieeetr}

\end{document}